\documentclass[11pt,logo,letterpaper]{berkeley}

\usepackage[utf8]{inputenc}
\usepackage[T1]{fontenc}
\usepackage{tgpagella}
\usepackage{microtype}

\usepackage{amsmath,amsthm,amsfonts}
\usepackage{mathrsfs}
\usepackage{bm}
\usepackage{latexsym}
\usepackage{bbding}
\usepackage{pifont}

\usepackage{marvosym}

\usepackage{xcolor}
\usepackage{graphicx}
\usepackage{subcaption}
\usepackage{tabularx}
\usepackage{adjustbox}
\usepackage{wrapfig}
\usepackage{float}
\usepackage{tikz}
\usetikzlibrary{shapes.geometric, arrows.meta, positioning, fit}
\usepackage[edges]{forest}
\usepackage{rotating}
\usepackage{eso-pic}
\usepackage{booktabs}
\usepackage{multirow}
\usepackage{array}
\usepackage{tabularx}
\usepackage{colortbl}
\usepackage{threeparttable}
\usepackage{longtable}
\usepackage{lscape}
\usepackage{makecell}

\usepackage{amssymb}

\definecolor{YaleBlue}{HTML}{2A5487}
\definecolor{CalGoldHex}{HTML}{FDF7F2}
\definecolor{TikTokPink}{HTML}{FF2B54}
\definecolor{IronGrey}{HTML}{6D6E71}
\definecolor{LinkPurple}{HTML}{FF2B54}
\definecolor{pipeline-orange}{RGB}{241,146,83}	
\definecolor{tree-pink}{RGB}{255,232,242}
\definecolor{tree-cyan}{RGB}{199,241,240}
\definecolor{tree-red}{RGB}{255,215,214}
\definecolor{tree-purple}{RGB}{230,231,255}
\definecolor{tree-green}{RGB}{192,242,213}
\definecolor{tree-yellow}{RGB}{255,242,230}
\definecolor{tree-blue}{RGB}{223,244,255}

\definecolor{box-pink}{RGB}{255,72,162}
\definecolor{box-cyan}{RGB}{29,234,221}
\definecolor{box-red}{RGB}{255,1,25}
\definecolor{box-purple}{RGB}{61,31,255}
\definecolor{box-green}{RGB}{71,212,90}
\definecolor{box-yellow}{RGB}{255,146,1}
\definecolor{box-blue}{RGB}{5,188,248}

\definecolor{hidden-draw}{RGB}{0,0,0}
\definecolor{hidden-pink}{RGB}{255,245,247}
\definecolor{hidden-red}{RGB}{205,44,36}
\definecolor{hidden-blue}{RGB}{194,232,247}
\definecolor{hidden-orange}{RGB}{243,202,120}
\definecolor{hidden-green}{RGB}{34,139,34}
\definecolor{hidden-black}{RGB}{20,68,106}
\definecolor{hidden-yellow}{RGB}{255,248,203}
\definecolor{purple}{RGB}{144,153,196}
\definecolor{yellow}{RGB}{255,228,123}
\definecolor{tkcolor}{RGB}{224,223,255}
\definecolor{level0}{rgb}{0.67, 0.88, 0.69}
\definecolor{level1}{rgb}{0.98, 0.92, 0.84}
\definecolor{level2}{rgb}{0.8, 0.8, 1.0}
\definecolor{level3}{rgb}{1.0, 0.71, 0.76}
\definecolor{level4}{rgb}{0.49, 0.99, 0.0}
\definecolor{level5}{rgb}{0.87, 0.63, 0.87}
\definecolor{darkblue}{rgb}{0, 0.40, 0.75}
\definecolor{mygreen}{RGB}{144, 238, 144}
\definecolor{darkmygreen}{RGB}{0, 100, 0}

\usepackage{caption}
\usepackage[round,authoryear]{natbib}
\usepackage{enumitem}
\setlist{
  itemsep=2pt,
  parsep=1pt,
  topsep=0pt,
  partopsep=0pt,
  leftmargin=*
}
\setlist[itemize]{itemsep=2pt, topsep=3pt, parsep=1pt}
\setlist[enumerate]{itemsep=3pt, topsep=4pt, parsep=1pt}
\setlist[description]{itemsep=3pt, topsep=4pt, parsep=1pt, style=nextline}
\usepackage{multicol}
\usepackage{titletoc}
\usepackage{titlesec}
\usepackage{algorithm,algpseudocode}
\usepackage{tcolorbox}
\usepackage{fontawesome}

\usepackage{hyperref}
\hypersetup{colorlinks=true, linkcolor=YaleBlue}
\usepackage{cleveref}

\tikzstyle{my-box}=[
rectangle,
draw=black,
rounded corners,
text opacity=1,
minimum height=1.5em,
minimum width=5em,
inner sep=2pt,
align=left,
fill opacity=.5,
]

\tikzstyle{section_2}=[my-box, fill=tree-pink]
\tikzstyle{section_3}=[my-box, fill=tree-blue]
\tikzstyle{section_4}=[my-box, fill=tree-cyan]
\tikzstyle{section_5}=[my-box, fill=tree-green]
\tikzstyle{section_6}=[my-box, fill=tree-purple]
\tikzstyle{appendix}=[my-box, fill=tree-yellow]

\tikzstyle{leaf}=[my-box, minimum height=1.5em, fill=tree-pink, text=black, align=left, font=\normalsize, inner xsep=5pt, inner ysep=4pt, text width=45em]
\tikzstyle{leaf2}=[my-box, minimum height=1.5em, fill=tree-cyan, text=black, align=left, font=\normalsize, inner xsep=5pt, inner ysep=4pt]
\tikzstyle{leaf3}=[my-box, minimum height=1.5em, fill=tree-red, text=black, align=left, font=\normalsize, inner xsep=5pt, inner ysep=4pt]
\tikzstyle{leaf4}=[my-box, minimum height=1.5em, fill=tree-purple, text=black, align=left, font=\normalsize, inner xsep=5pt, inner ysep=4pt]

\newlength{\sectionbeforeskip}
\newlength{\sectionafterskip}
\newlength{\subsectionbeforeskip}
\newlength{\subsectionafterskip}
\newlength{\subsubsectionbeforeskip}
\newlength{\subsubsectionafterskip}
\newlength{\openingafterspace}
\newlength{\epigraphafterspace}
\newlength{\leadparagraphskip}
\newlength{\guidingquestionaboveskip}
\newlength{\guidingquestionbelowskip}
\newlength{\sectionsubtitleaboveskip}
\newlength{\sectionsubtitlebelowskip}
\newlength{\remarkboxskip}
\newlength{\tocheaderskip}
\newlength{\tocrulesep}
\newlength{\tocbodyskip}

\titlespacing*{\section}{0pt}{\sectionbeforeskip}{\sectionafterskip}
\titlespacing*{\subsection}{0pt}{\subsectionbeforeskip}{\subsectionafterskip}
\titlespacing*{\subsubsection}{0pt}{\subsubsectionbeforeskip}{\subsubsectionafterskip}

\newcommand{\openingstatement}[1]{%
  \begin{quote}
    \color{YaleBlue}\itshape #1
  \end{quote}
  \vspace{\openingafterspace}
}

\newcommand{\chapterepigraph}[3]{%
  \begin{quote}
    \color{YaleBlue}\itshape
    ``#1''\\[0.8ex]
    {\normalfont\small\hspace*{\fill}--- \textsc{#2}%
    \if\relax\detokenize{#3}\relax\else\\
      \hspace*{\fill}\footnotesize\textit{#3}%
    \fi}
  \end{quote}
  \vspace{\epigraphafterspace}
}

\newcommand{\leadparagraph}[1]{%
  \par\vspace{\leadparagraphskip}\noindent\textbf{#1}\enspace
}

\newcommand{\labeledlead}[2]{%
  \leadparagraph{#1. #2}
}

\newcommand{\guidingquestion}[1]{%
  \par\vspace{\guidingquestionaboveskip}%
  \noindent\textcolor{IronGrey}{\itshape #1}%
  \par\vspace{\guidingquestionbelowskip}
}

\newcommand{\sectionsubtitle}[1]{%
  \par\vspace{\sectionsubtitleaboveskip}%
  \noindent{\color{IronGrey}\itshape #1}%
  \par\vspace{\sectionsubtitlebelowskip}
}

\newcommand{\rendercondensedtoc}{%
  \addtocontents{toc}{\protect\setcounter{tocdepth}{2}}%
  \vspace*{\tocheaderskip}%
  \startcontents[sections]\vbox{\sc Table of Contents}%
  \vspace{\tocrulesep}%
  \hrule height .8pt
  \vspace{-2mm}
  {\setlength{\baselineskip}{11pt}%
  \setlength{\parskip}{3pt}%
  \printcontents[sections]{l}{1}{\setcounter{tocdepth}{2}}}%
  \vspace{\tocrulesep}%
  \hrule height .8pt
  \vspace*{\tocbodyskip}%
}

\tcbset{
  takeawaysbox/.style={
    colback=CalGoldHex,
    colframe=YaleBlue,
    coltext=IronGrey,
    fontupper=\footnotesize,
    width=\linewidth,
    arc=2mm,
    boxrule=0.2mm,
    top=4mm,
    bottom=3mm,
    left=5mm,
    right=5mm,
    before skip=\remarkboxskip,
    after skip=\remarkboxskip,
  }
}

\newtcolorbox[auto counter]{remarkboxenv}[2][]{
  takeawaysbox,
  colbacktitle=YaleBlue,
  coltitle=white,
  fonttitle=\bfseries,
  title={Remark \thetcbcounter: #2},
  #1
}

\tcbuselibrary{breakable}
\newcommand{\remarkbox}[1]{
\begin{tcolorbox}[
    colback=gray!10, 
    colframe=gray!50, 
    coltitle=black, 
    width=\textwidth, 
    boxrule=0.5pt, 
    arc=1mm, 
    boxsep=1mm,
    left=3mm,
    right=3mm,
    top=3mm,
    bottom=3mm,
    before skip=4mm,
    after skip=4mm,
    breakable
]
\textbf{Remarks}. {#1}
\end{tcolorbox}
}

\title{AgentSchool: An LLM-Powered Multi-Agent Simulation for Education}
\runningtitle{AgentSchool: An LLM-Powered Multi-Agent Simulation for Education}

\author{
\textbf{Yulei Ye}\textsuperscript{\rm 1,$\dag$}
\hfill \textbf{Wenhao Li}\textsuperscript{\rm 2,$\dag$}
\hfill \textbf{Zhong Wen}\textsuperscript{\rm 3}
\hfill \textbf{Yunshu Huang}\textsuperscript{\rm 3}
\hfill \textbf{Yichen Hu}\textsuperscript{\rm 3}
\hfill \textbf{Zifan Wei}\textsuperscript{\rm 4} 
\textbf{Yige Wang}\textsuperscript{\rm 1} 
\hfill \textbf{Xinyu Xie}\textsuperscript{\rm 3} 
\hfill \textbf{Haoxuan Yang}\textsuperscript{\rm 5}
\hfill \textbf{Yanjun Huang}\textsuperscript{\rm 1}
\hfill \textbf{Ruijia Li}\textsuperscript{\rm 5}
\hfill \textbf{Hong Qian}\textsuperscript{\rm 1} 
\textbf{Yu Song}\textsuperscript{\rm 1}
\hfill \textbf{Bo Jiang}\textsuperscript{\rm 1}
\hfill \textbf{Bingdong Li}\textsuperscript{\rm 1}
\hfill \textbf{Lijun Li}\textsuperscript{\rm 6}
\hfill \textbf{Bo Zhang}\textsuperscript{\rm 6} 
\hfill \textbf{Pinlong Cai}\textsuperscript{\rm 6}
\textbf{Xingcheng Xu}\textsuperscript{\rm 6}
\hfill \textbf{Shuangye Chen}\textsuperscript{\rm 5}
\hfill \textbf{Xia Hu}\textsuperscript{\rm 6}
\hfill \textbf{Liang He}\textsuperscript{\rm 3}
\textbf{Aimin Zhou}\textsuperscript{\rm 1} \\
\textbf{Jingjing Qu}\textsuperscript{\rm 6,$\ddag$}
\textbf{Jing Shao}\textsuperscript{\rm 6,$\ddag$}
\textbf{Xiangfeng Wang}\textsuperscript{\rm 1,$\ddag$} \\
\vspace{2mm}
\textsuperscript{\rm 1} Shanghai Institute of AI for Education, East China Normal University \\
\textsuperscript{\rm 2} School of Computer Science and Technology, Tongji University \\
\textsuperscript{\rm 3} School of Computer Science and Technology, East China Normal University \\
\textsuperscript{\rm 4} School of Design, East China Normal University \quad
\textsuperscript{\rm 5} Faculty of Education, East China Normal University \quad
\textsuperscript{\rm 6} Shanghai Artificial Intelligence Laboratory\\
\vspace{0.8mm}
{\footnotesize \textsuperscript{$\dag$}Equal contribution. \quad \textsuperscript{$\ddag$}Corresponding author.}
}
\correspondingauthor={}

\begin{document}

\begin{abstract}
Despite the rapid deployment of large language models (LLMs) into classrooms, validating educational AI remains uniquely intractable: interventions act on developing learners whose cognitive and social trajectories are irreversibly shaped, while real-world trials are slow, ethically constrained, and institutionally locked. 
LLM-based educational simulators have emerged as a potential remedy, but many still collapse learning into persona-conditioned role-play and, when optimized only to reproduce existing classrooms, can structurally penalize the institutional novelty that pedagogical reform requires. 
In this work, we bridge this gap by introducing AgentSchool, an LLM-driven multi-agent simulator that models learning as state transition rather than prompted behavior. 
AgentSchool couples cognitively growable student agents—equipped with weighted subject knowledge graphs, thinking-workflow pools, and explicit misconceptions—with adaptive teacher agents that plan, scaffold, and reflect along the Zone of Proximal Development, embedded in a configurable scenery generator that situates instruction within both formal and informal learning fields, and a multi-scale simulator that decouples interaction scale, temporal granularity, and simulation duration. 
In a $2\times3$ controlled lesson study across five backbone LLMs, structured student agents produce more differentiated mastery and misconception traces than a baseline simulator, while teacher-agent comparisons show backbone-dependent patterns consistent with ZPD-informed adaptation.
In informal social scenes, AgentSchool generates plausible traces of peripheral participation, clique formation, aggressor-induced cohesion, and opinion-leader emergence consistent with classroom social theories. 
Beyond its role as an educational research instrument, AgentSchool frames education as a socially meaningful testbed for long-horizon memory, heterogeneous multi-agent coordination, and future institutional reasoning under organizational pressure.

\par\vspace{0.35em}
\centering
\includegraphics[width=0.9\linewidth]{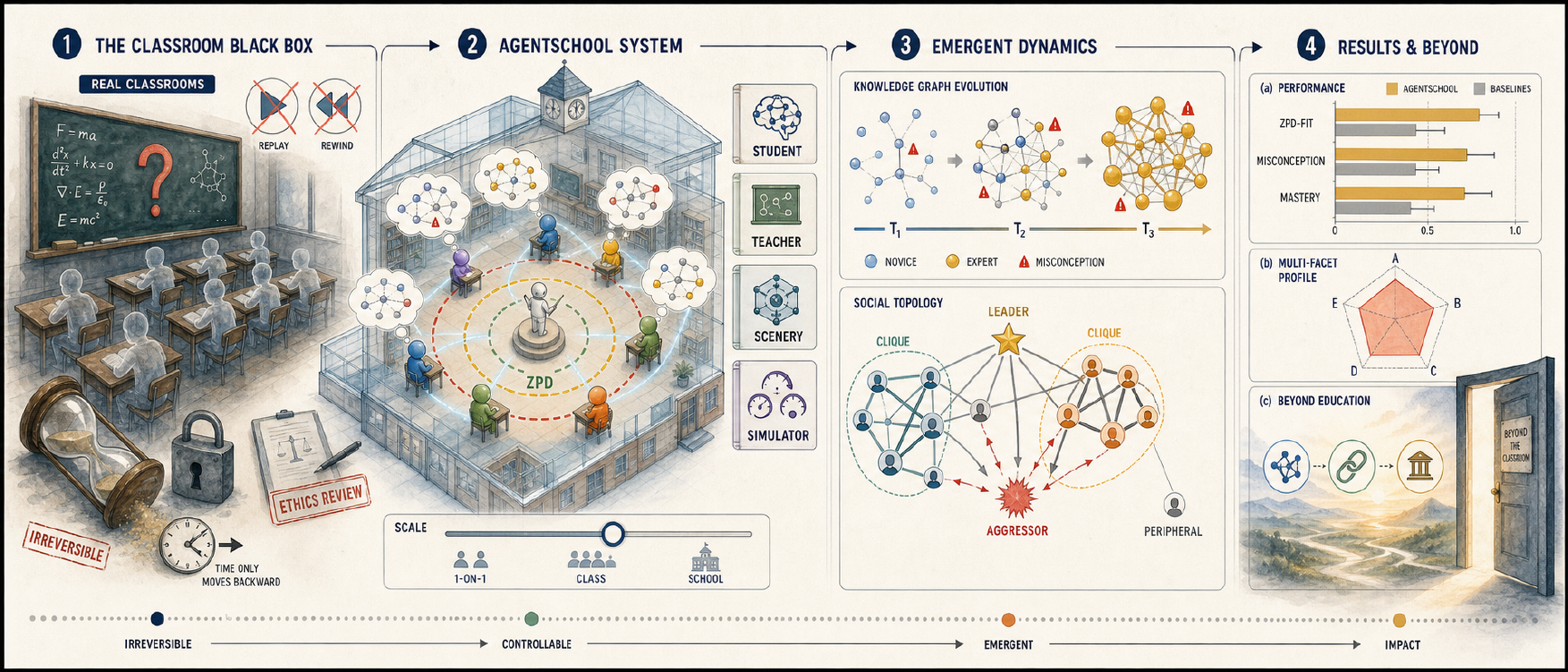}
\par\vspace{0.2em}
{\small Overview of AgentSchool.}
\label{fig:paper_overview}

\end{abstract}

\AddToShipoutPictureBG*{
  \AtPageUpperLeft{
    \put(\LenToUnit{2.0cm},\LenToUnit{-2.05cm}){
      \includegraphics[width=3.6cm, height=0.7cm, keepaspectratio]{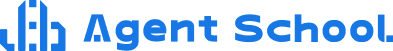}
    }
  }
}
\AddToShipoutPictureBG*{
  \AtPageUpperLeft{
    \put(\LenToUnit{6.1cm},\LenToUnit{-2.12cm}){
      \includegraphics[width=4.0cm, height=0.75cm, keepaspectratio]{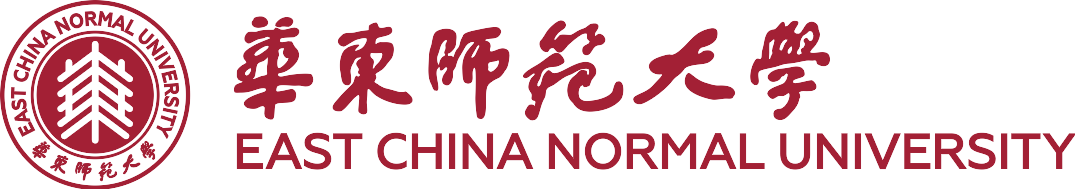}
    }
  }
}
\AddToShipoutPictureBG*{
  \AtPageUpperLeft{
    \put(\LenToUnit{10.7cm},\LenToUnit{-2.20cm}){
      \includegraphics[width=2.5cm, height=0.83cm, keepaspectratio]{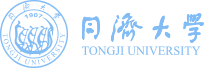}
    }
  }
}
\AddToShipoutPictureBG*{
  \AtPageUpperLeft{
    \put(\LenToUnit{13.9cm},\LenToUnit{-2.05cm}){
      \includegraphics[width=5.2cm, height=0.78cm, keepaspectratio]{logo-shlab-horizontal.png}
    }
  }
}

\maketitle

\clearpage
\vspace*{-1cm}
\rendercondensedtoc

\newcommand{\fix}{\marginpar{FIX}}
\newcommand{\new}{\marginpar{NEW}}

\openingstatement{AgentSchool is motivated by a simple but high-stakes question: before educational AI is deployed into real classrooms, can we rehearse its consequences in a plausible, inspectable, and ethically safer simulation environment? We present AgentSchool as a computational ``wind tunnel'' for future education, where students, teachers, scenarios, and eventually institutions can be modeled as evolving agents rather than static roles.}

\section{Introduction}\label{sec:intro}
\guidingquestion{How can researchers evaluate future educational systems when real-world trials are slow, institutionally constrained, and ethically irreversible?}

\labeledlead{The Shock}{Generative AI (GenAI), and large language models (LLMs) in particular, are destabilizing the assumptions behind modern schooling.}
Legacy assessment systems often measure lower-order thinking skills (LOTS), precisely the skills that GenAI can now automate at scale~\citep{Chavda2023, Mhlanga2024}. This creates an assessment crisis: product-based evaluation is no longer a reliable proxy for the learning process~\citep{Bozkurt2024, Smith2024}. When GenAI functions as an ``automated contract cheating'' tool~\citep{Lancaster2024}, student output can be separated from student understanding. Empirical evidence further suggests that passive, answer-seeking AI use may weaken critical thinking, whereas constructive and knowledge-building use can support deeper learning~\citep{Cela2024, Pallant2025}.

The educational task therefore shifts from transmitting answers to cultivating capacities that remain distinctly human: evaluative judgement, creativity, and higher-order thinking~\citep{Bearman2024}. Yet institutional reform moves more slowly than technological change~\citep{Wang2021}. Without deliberate governance, AI adoption is likely to be driven by efficiency rather than equity~\citep{AlEmran2025}, potentially widening the digital divide~\citep{Jefferson2024} and embedding algorithmic bias~\citep{Bozkurt2024b, Haleem2024}. This urgency has already shaped major policy agendas, including China's Ministry of Education \textit{White Paper on Smart Education}, which frames 2025 as the inaugural year of smart education and envisions AI-integrated teachers, classrooms, schools, and learning centers as central pillars of future education~\citep{MEC_smarteducationwhitebook}.

This tension creates a methodological problem. Educational AI is not merely another digital tool whose performance can be judged by accuracy, latency, or user satisfaction. It participates in the formation of learners' habits of attention, epistemic trust, social identity, and long-term relationship with knowledge. A recommendation engine may be evaluated after deployment by measuring click-through or retention; an educational intervention cannot be treated in the same way, because the object being optimized is a developing person rather than an already-formed consumer preference. A system that improves short-term answer correctness may still erode persistence, curiosity, or the ability to reason without assistance. Conversely, an intervention that looks inefficient in a short trial may cultivate durable capacities that only become visible later. Any validation method for educational AI must therefore represent both immediate performance and developmental trajectory.

\begin{remarkboxenv}{Why Educational AI Needs a Wind Tunnel}
Educational AI cannot be validated only by functional accuracy. It acts on children and adolescents whose learning habits, reasoning patterns, social relationships, and self-concepts are still forming. A harmful intervention may not be reversible after deployment. AgentSchool is designed to make the consequences of educational AI visible before those consequences reach real learners.
\end{remarkboxenv}

\labeledlead{The Barrier}{Traditional reform pathways are constrained by institutional inertia.}
At the institutional level, school systems exhibit strong \textit{path dependency}: high-stakes standardized testing, performativity cultures, and staff-based budgeting models lock systems into existing routines and suppress pedagogical innovation~\citep{Hagood2021, Champion2016, Muvunyi2024, Roza2010}. At the cultural level, teaching is itself a ``cultural activity''~\citep{Stigler2015}. Practitioners must unlearn deeply embedded beliefs and classroom scripts before new approaches can take hold~\citep{Doering2002}. Even successful local innovations often remain person-dependent and fail to reach the critical mass needed for systemic change~\citep{Tulyakul2023, Sotiriou2018}.

These structural constraints are intensified by the high stakes of educational AI. Unlike many application domains, education intervenes during developmental periods in which learners' cognitive, emotional, and social capacities are highly malleable. Ethical harms cannot simply be repaired after the fact~\citep{Floridi2018}. Large-scale AI deployment in education therefore requires validation protocols that evaluate not only technical reliability, but also learning outcomes, developmental well-being, and institutional equity.

The difficulty is that these outcomes are coupled. A change in assessment design may reshape teacher workload; teacher workload may alter feedback quality; feedback quality may change student motivation; student motivation may then affect peer culture and classroom climate. These are not independent variables that can be isolated cleanly in a conventional laboratory study. They are feedback processes unfolding across different temporal scales. A single lesson may reveal whether an AI tutor gives correct explanations, but it cannot reveal whether repeated use changes student agency. A semester-long randomized trial may reveal average learning gains, but it is expensive, ethically burdensome, and usually too slow to guide a fast-moving technology ecosystem. We therefore need a complementary method that can explore plausible causal pathways before full-scale field deployment.

\labeledlead{The Opportunity}{The same models that disrupt education also make a new validation methodology possible.}
LLM-based social simulation (LSS), also known as LLM-agent-based modeling (LLM-ABM), can generate high-fidelity sandboxes for complex social behavior~\citep{Park2023, An2021, piao2025agentsociety}. Micro-level simulations such as \texttt{SimClass}~\citep{Zhang2025} and \texttt{EduVerse} test interventions within classroom dynamics. Macro-level simulations have begun to anticipate emergent policy consequences, such as predicting that an education voucher program would reduce enrollment among disadvantaged students~\citep{Anon2024_PolicySim}. This points toward a shift from reactive to anticipatory policy-making.

Existing educational simulations, however, still face two structural limitations:
\begin{itemize}
    \item \textbf{Fidelity gap.} Learning is simultaneously cognitive, emotional, social, and institutional. Prompt-engineering alone cannot represent the process-level dynamics through which outcomes emerge over time.
    \item \textbf{Reliability gap.} LLM reasoning is opaque, and most simulation systems do not support long-term data collection, human-in-the-loop inspection, or multi-level evaluation mechanisms.
\end{itemize}

These limitations are especially serious because many current systems are optimized for reproducing existing classroom behavior rather than exploring new institutional configurations~\citep{Zhao2024d}. A simulator that merely rewards fidelity to the present can penalize the very novelty required to overcome systemic inertia. It also struggles to model supra-rational organizational behaviors, including the pursuit of institutional legitimacy, that drive path-dependent lock-in~\citep{DiMaggio1983}.

We distinguish three validation targets that are often conflated. \textit{Behavioral believability} asks whether generated dialogue resembles what teachers and students might say. \textit{Educational mechanism fidelity} asks whether the internal process producing that dialogue corresponds to learning theory, such as knowledge restructuring, scaffolding, or misconception repair. \textit{Institutional counterfactual usefulness} asks whether the simulator can support ``what-if'' reasoning about policies, spaces, or AI-mediated teaching arrangements that do not yet exist. A classroom role-play system can satisfy the first target while failing the second and third. AgentSchool is designed around the latter two targets: it makes internal learning states explicit, records state transitions, and allows scenarios to depart from present-day classroom templates in a controlled manner.

\labeledlead{Our Response}{We introduce \textbf{AgentSchool}\footnote{\url{https://github.com/epitome-AISS/AgentSchool}}, an LLM-driven multi-agent simulation platform for modeling educational ecosystems.}
AgentSchool operationalizes teaching and learning within a parametrically controlled virtual environment grounded in educational theory and empirical research, while exposing interfaces that can later support management, assessment, and policy-level simulation. It is organized around four design commitments:
\begin{enumerate}[label=\textbf{C\arabic*.}, leftmargin=*]
    \item \textbf{Scenario configurability.} Modular scenario templates represent classroom conditions and are designed to extend toward institutional pressures such as mimetic isomorphism~\citep{DiMaggio1983}, legitimacy seeking, and imitation of high-status peers.
    \item \textbf{Closed instructional loop with extensible governance.} The current platform represents the teach--learn--assess loop within configurable learning fields, with management and evaluator roles treated as planned extensions for institutional-level analysis.
    \item \textbf{Empirical calibration.} Agent behaviors can be grounded in real-world educational data where available, while theory-derived constraints provide a disciplined baseline when such data are scarce.
    \item \textbf{Temporal fast-forwarding.} The simulator is designed to support long-horizon analysis of how early design decisions may trigger self-reinforcing feedback loops; the present paper examines this agenda through shorter formal and informal simulations.
\end{enumerate}

\paragraph{Terminology.}
We use \textit{learning field} for the theoretical construct: the sociocultural, material, relational, and temporal field in which learning occurs. We use \textit{scenario} for an experimental specification of that field, including its participants, resources, rules, and activity structure. We use \textit{scenery} for the implemented module that instantiates and updates such scenarios inside AgentSchool.

Formally, AgentSchool models an educational system as a partially observable, multi-agent state transition process. Let $\mathcal{N}=\mathcal{S}\cup\mathcal{T}\cup\mathcal{O}$ denote the set of student agents, teacher agents, and organizational or environmental agents. At simulation step $t$, the system state is
\[
    X_t = \left(\{x^s_{i,t}\}_{i=1}^{|\mathcal{S}|}, \{x^t_{j,t}\}_{j=1}^{|\mathcal{T}|}, C_t, G_t, H_t \right),
\]
where $x^s_{i,t}$ is the internal state of student $i$, $x^t_{j,t}$ is the state of teacher $j$, $C_t$ is the current scenery configuration, $G_t$ is the social and instructional interaction graph, and $H_t$ is the accumulated history of events. Agents choose actions from local observations $o_{i,t}=\Omega_i(X_t)$ rather than from the full state, and the simulator applies a transition operator
\[
    X_{t+1} \sim \mathcal{P}_{\theta}(X_t, a_{1,t},\ldots,a_{n,t}),
\]
where $\theta$ includes model parameters, scenario parameters, pedagogical rules, and calibration settings. This formulation makes explicit that AgentSchool is not a static prompt collection; it is a mechanism for generating, inspecting, and comparing educational trajectories.

Our contribution is therefore threefold. First, we propose a cognitively growable student-agent architecture that represents knowledge mastery, reasoning workflows, learning modality, memory, and misconceptions as mutable state variables. Second, we design adaptive teacher agents that plan, scaffold, observe, and reflect according to student states and ZPD-aligned instructional objectives. Third, we integrate these agents into a scenery-driven simulator that supports formal teaching, informal socialization, and future-oriented scenario exploration. Together, these components allow AgentSchool to evaluate not only whether an educational interaction looks realistic, but also whether the underlying trajectory is educationally interpretable.

\begin{figure}[htb!]
\centering
\includegraphics[width=\textwidth]{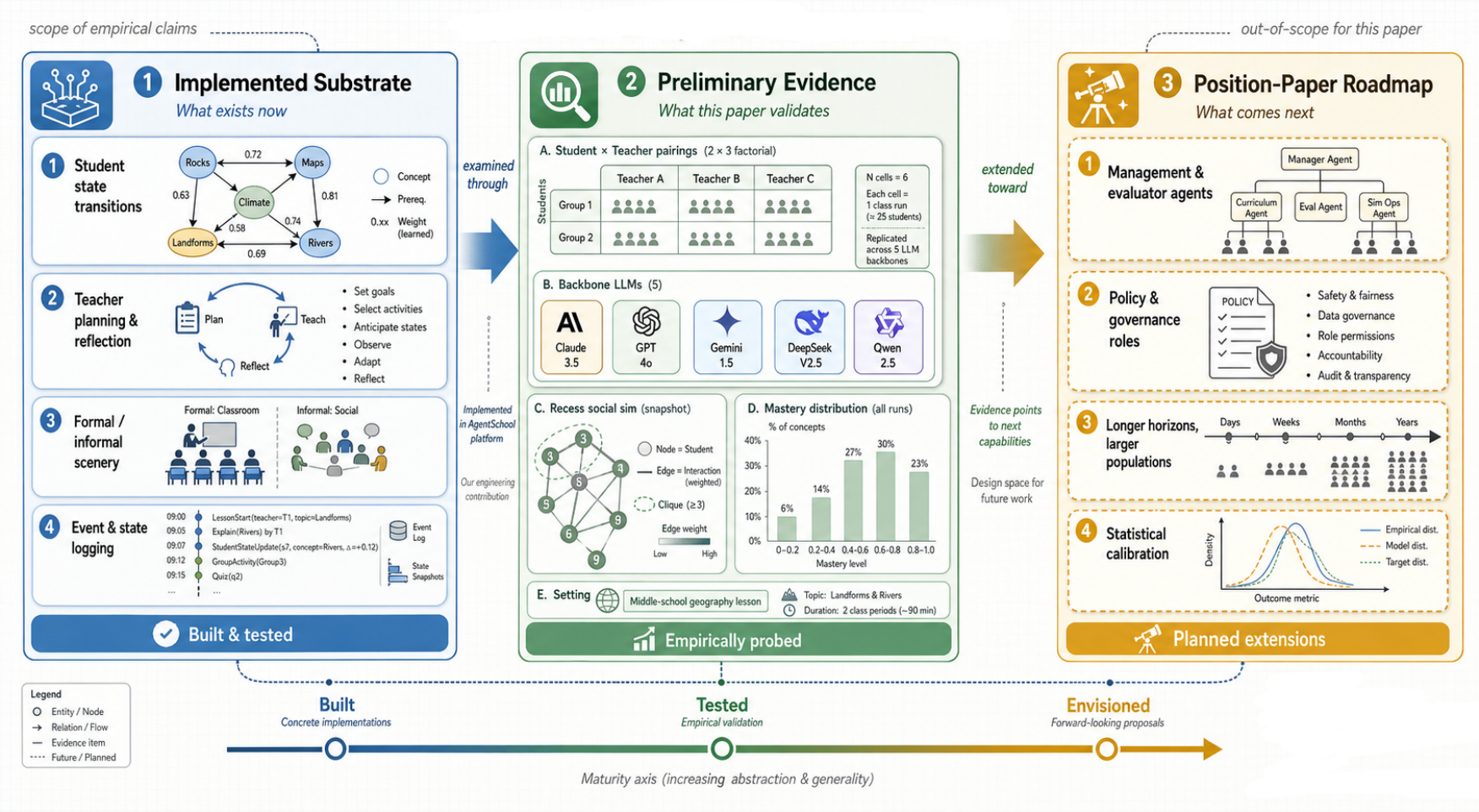}
\caption{Scope boundary of the present paper. AgentSchool's implemented substrate is examined through preliminary lesson and social simulations; institutional and policy-level uses are positioned as extensions rather than completed empirical claims.}
\label{fig:scope_boundary}
\end{figure}

\remarkbox{In short, AgentSchool is both an education-research instrument and an AI-research testbed. For educators and policymakers, it offers a safer space for prototyping future educational models. For the agent community, it provides a socially meaningful environment for studying long-term memory, multi-agent coordination, and agentic reasoning.}

\section{Related Work}
\label{sec:relatedwork}
\sectionsubtitle{We review prior work through four lenses: educational scenarios, agent-based simulation, LLM-enabled simulation technology, and simulator calibration. The purpose is not only to summarize existing systems, but to identify what an educational ``wind tunnel'' must be able to represent.}

\subsection{Educational Scenario}

\guidingquestion{What counts as an educational scenario if learning is not confined to a classroom?}

Diverse learning scenarios reflect how learning itself is conceptualized~\citep{fischer_internationalhandbooklearning_2018}. A scenario is not a static room. It is a dynamic, meaning-laden learning field composed of people, resources, spatial arrangements, tools, schedules, norms, and interaction patterns. It provides both possibilities and constraints: what can be said, who can participate, how knowledge circulates, and how quickly roles can change.

For the purpose of simulation, we define an educational scenario as
\[
    C = \left(\mathcal{N}^{C}, \mathcal{L}^{C}, \mathcal{Q}^{C}, \mathcal{U}^{C}, \mathcal{B}^{C}, \tau^{C} \right),
\]
where $\mathcal{N}^{C}$ denotes participating agents, $\mathcal{L}^{C}$ denotes their social and pedagogical relations, $\mathcal{Q}^{C}$ denotes material and symbolic resources, $\mathcal{U}^{C}$ denotes permissible activities, $\mathcal{B}^{C}$ denotes norms and constraints, and $\tau^{C}$ denotes the temporal rhythm of the setting. This definition is intentionally broader than ``classroom layout.'' A lecture hall, a group-discussion room, a learning-management system, a peer chat during recess, and a future AI-supported learning center may have different values for all six components even when they share the same curriculum content. If a simulator ignores any of these dimensions, it risks attributing outcomes to the wrong cause: a student may fail not because of low ability, but because the scenario affords only passive response; a group may become productive not because of superior content, but because the spatial and normative structure supports mutual explanation.

\labeledlead{Pedagogical structure}{Scenarios define how knowledge flows and what roles participants assume.}
Lecture-based classrooms often follow the closed Initiate--Respond--Evaluate (IRE) pattern~\citep{barr_learninglessonssocial_1980}. By contrast, intellectually engaged classrooms rely on dialogue, group discussion, argumentative reasoning, and presentation, positioning students as thinkers and reasoners. Computer-supported collaborative learning further operationalizes these patterns through scaffolds and scripts~\citep{dillenbourg_mechanicscsclmacro_2008}, including argumentation environments~\citep{tsovaltzi_groupawarenesssupport_2014} and peer-feedback templates~\citep{gielen_scriptingroleassessor_2015}.

The pedagogical structure of a scenario determines the admissible discourse moves. In an IRE sequence, the teacher typically controls topic initiation and evaluation, while students provide short responses. In collaborative inquiry, students can initiate claims, request evidence, challenge assumptions, and build shared artifacts. In peer feedback, a learner alternates between producer, assessor, and reviser. These structures imply different transition probabilities over actions. A role-play simulator that gives every agent the same conversational affordances will miss the institutional asymmetry of a classroom; a simulator that fixes the teacher as the only legitimate initiator will miss the possibility of learner agency. AgentSchool therefore treats pedagogical structure as an explicit scenario parameter rather than as background prose in the prompt.

\labeledlead{Material structure}{Physical and technological environments carry pedagogy while reshaping behavior.}
Rows of fixed desks reinforce hierarchical, teacher-centered interaction. Round tables, group workstations, reading corners, laboratories, and digital whiteboards can instead encourage peer exchange, experimentation, and shared representation. The same instructional objective may therefore produce different social dynamics under different spatial and technological affordances.

Material structure matters because educational interaction is embodied and tool-mediated. A student seated at the back of a fixed-row classroom has a different opportunity set from a student sitting in a four-person group with shared manipulatives. A collaborative whiteboard changes not only where content is displayed, but also who can modify it and how revisions become visible. Online platforms add further constraints: asynchronous forums support delayed reflection but reduce immediate affective cues; synchronous video rooms allow turn-taking but may amplify silence; intelligent tutoring systems personalize sequence but may narrow the social field. In AgentSchool, these material conditions are represented as constraints and affordances attached to the scenery, so that interaction patterns can be traced back to the environment that produced them.

\labeledlead{Temporal structure}{School life is a sequence of formal, informal, and semi-structured situations.}
Formal classes sit alongside breaks, morning reading, independent study, club activities, and tutoring. Informal moments are not empty time: they shape peer norms, identity, belonging, and the classroom climate. Thornburg's four metaphors--the campfire, watering hole, cave, and life--capture the need to balance expert instruction, collaborative exchange, reflection, and real-world application~\citep{thornburg_campfireholodeckcreating_2013}.

Temporal structure is particularly important for future educational simulation because learning rarely occurs at a single granularity. A misconception may appear in one utterance, persist across a lesson, and be repaired only after repeated encounters with conflicting evidence. A peer relationship may begin during a break, affect group participation during class, and later influence academic confidence. Thus, a scenario should not be modeled as an isolated snapshot; it should be modeled as part of a schedule and a history. AgentSchool's scenery generator represents this by allowing scenarios to be composed into sequences, so that formal instruction and informal social experience can jointly influence the learner state.

\remarkbox{For AgentSchool, a scenario is therefore a configurable relational field. It must include agents, resources, space, time, rules, and latent social meaning, because educational outcomes emerge from their interaction rather than from instructional content alone.}

\subsection{Agent-Based Modeling and Simulation in Various Fields}

Agent-Based Modeling and Simulation (ABMS) models a system by representing its parts as agents and its mechanisms as interactions among those agents. It has been widely used in education~\citep{gu_systematicreviewagentbased_2015,bodine_agentbasedmodelingsimulation_2020}, medicine~\citep{li_agenthospitalsimulacrum_2025,hicke_medsimaisimulationformative_2025}, economics~\citep{xie_canlargelanguage_2024}, and sociology~\citep{mi_mfllmsimulatingpopulation_2025}.

The core epistemic promise of ABMS is that macro-level regularities can be explained as the cumulative result of micro-level interaction. In a conventional equation-based model, the researcher often specifies aggregate dynamics directly. In ABMS, the researcher specifies agent attributes, local rules, and interaction topology, then observes whether system-level patterns emerge. This is especially suitable for education because many educational phenomena are emergent: classroom climate arises from repeated teacher-student and peer interactions; achievement gaps are reinforced by feedback loops among resources, expectation, motivation, and opportunity; institutional routines persist because individual actors adapt rationally to local constraints even when the aggregate outcome is undesirable.

Mathematically, an ABMS can be described as a tuple
\[
    \mathcal{M}_{\mathrm{ABMS}} = \left(\mathcal{N}^{A}, \mathcal{E}^{A}, \mathcal{I}^{A}, \mathcal{U}^{A}, \mathcal{Y}^{A}\right),
\]
where $\mathcal{N}^{A}$ is the agent set, $\mathcal{E}^{A}$ is the environment, $\mathcal{I}^{A}$ is the interaction topology, $\mathcal{U}^{A}$ is the set of update rules or policies, and $\mathcal{Y}^{A}$ is the set of observable outputs. AgentSchool extends this classical structure by replacing many hand-coded update rules with LLM-supported decision modules while retaining explicit state variables and logging mechanisms. This hybridization is important: LLMs supply linguistic and contextual flexibility, but educational validity still requires structured states, constraints, and evaluation variables.

\begin{table}[!htbp]
\centering
\caption{How ABMS informs educational simulation.}
\label{tab:abms_landscape}
\renewcommand{\arraystretch}{1.05}
\begin{tabularx}{\linewidth}{@{}p{0.18\linewidth}X X@{}}
\toprule
\textbf{Field} & \textbf{Typical focus} & \textbf{Value for AgentSchool} \\
\midrule
Education & Students, teachers, school leaders, learning outcomes & Models instructional interaction and learner development~\citep{dubovi_instructionalsupportlearning_2019}. \\
Sociology & Group dynamics, diffusion, norm formation & Supports macro-level policy and social-network analysis~\citep{mi_mfllmsimulatingpopulation_2025}. \\
Economics & Individual decision-making, trust, incentives & Provides methods for micro-level behavioral alignment~\citep{xie_canlargelanguage_2024}. \\
Medicine & Patient simulation, vocational skill practice & Offers analogues for high-stakes professional training~\citep{hicke_medsimaisimulationformative_2025}. \\
\bottomrule
\end{tabularx}
\end{table}

Educational simulation follows two main motivations. The first is \textbf{simulation for educational research}, which explores new methods, patterns, and laws~\citep{dubovi_instructionalsupportlearning_2019}. Recent work predicts student answers~\citep{xu_classroomsimulacrabuilding_2025} or generates teacher--student interaction to improve teaching plans~\citep{hu_exploringpotentialllm_2025}. The second is \textbf{simulation for education}, where simulation itself becomes a learning tool~\citep{bhowmik_evaluationllmpoweredstudent_2024}; examples include situated learning supports~\citep{pan_tutorupwhatif_2025}, supplemental materials~\citep{takahashi_datascienceagentbased_2023}, learning-channel expansion~\citep{stummer_agentbasedmarketsimulation_2021}, ecological modeling scaffolds~\citep{basu_scaffoldingframeworksupport_2015}, and simulated medical-patient tutoring~\citep{hicke_medsimaisimulationformative_2025}.

These two motivations impose different evaluation standards. If simulation is used as an instructional activity, the primary question is whether learners benefit from interacting with the simulated environment. If simulation is used as a research instrument, the primary question is whether the simulated environment can support valid inference about educational mechanisms. AgentSchool belongs primarily to the second category, although its interface can support exploratory use by practitioners. This distinction matters because a system can be engaging for learners while still being too under-specified for research claims. Conversely, a research simulator may be less playful but more useful for examining causal structure, boundary conditions, and long-horizon consequences.

Existing educational ABMS also tends to focus on one layer at a time. Classroom models often provide detailed teacher-student interaction but do not represent school governance, policy pressure, or informal peer networks. Policy models often represent school choice or resource allocation but simplify classroom learning into aggregate outcome variables. AgentSchool aims to connect these layers through a shared state-transition representation. A lesson can update individual knowledge states; repeated lessons and social interactions can update classroom networks; network-level effects can then feed back into participation, teacher strategy, and institutional indicators. This vertical coupling is necessary for studying educational AI because AI tools may intervene simultaneously at the level of tutoring, assessment, teacher planning, administration, and policy.

\begin{remarkboxenv}{Research Gap}
Existing educational simulations remain strong at local classroom interaction but weak at institutional modeling. Management, policy, and school-level social dynamics are often omitted, even though policy effects propagate from macro structures to micro classroom practice~\citep{maroulis_complexsystemsview_2010, maroulis_modelingtransitionpublic_2014}.
\end{remarkboxenv}

\subsection{Technical Factors in Agent-Based Modeling and Simulation}

\guidingquestion{What technical trade-offs determine whether an LLM-based simulation is useful rather than merely expressive?}

Before LLMs, ABMS often relied on predefined rules and functions. Such systems can generate complex aggregate behavior from simple local rules, as in FTS-SOCI for simulating teaching strategies and sociogram evolution~\citep{garcia-magarino_ftssociagentbasedframework_2015}. Yet rule-defined agents typically lack the linguistic flexibility and individual diversity needed for rich educational interaction.

LLM-based agents changed this design space. Park et al. showed that agents equipped with observation, planning, and reflection could produce believable individual and emergent social behaviors~\citep{park_generativeagentsinteractive_2023}. Subsequent systems, such as Agent Hospital, extended this approach to complex professional domains by using autonomous agents as patients, nurses, and doctors~\citep{li_agenthospitalsimulacrum_2025}.

The technical challenge is that LLM-based agents are both powerful and unstable. They can generate context-sensitive utterances, infer intentions, and adapt to nuanced social situations. However, without architectural constraints they can also drift from their assigned role, invent inconsistent memories, overfit to prompt wording, or produce surface-level plausibility without maintaining a coherent internal state. Educational simulation magnifies these problems because the target behavior is not merely conversation; it is learning. A student agent should not only say plausible student-like sentences, but also exhibit constrained growth: prior knowledge should shape interpretation, misconceptions should persist until challenged, and mastery should change gradually rather than randomly. A teacher agent should not merely be friendly or verbose, but should diagnose learner state, choose scaffolds, and adjust task difficulty.

\begin{remarkboxenv}{The Simulation Triangle}
LLM-based simulation is governed by three coupled variables: \textbf{scale}, \textbf{duration}, and \textbf{granularity}. Increasing one usually pressures the others. A useful educational simulator must make this trade-off configurable rather than fixed.
\end{remarkboxenv}

We can formalize this trade-off by representing a simulation budget as
\[
    B \approx N \cdot H \cdot \frac{1}{\Delta t} \cdot c_{\mathrm{step}},
\]
where $N$ is the number of active agents, $H$ is the simulated horizon, $\Delta t$ is the temporal granularity of each step, and $c_{\mathrm{step}}$ is the average computational cost per agent-step. Fine-grained dialogue lowers $\Delta t$ and increases cost; large institutional populations increase $N$; long-term development increases $H$. Because $B$ is finite, every simulator makes an implicit design decision about which educational mechanisms remain visible. AgentSchool exposes this decision by allowing researchers to choose the granularity appropriate to the research question.

\labeledlead{Scale}{Simulations range from single-agent studies to planetary-scale social systems.}
When agent populations exceed 1,000, computational cost and response latency become central barriers. Piao et al. generated more than 10,000 agents and simulated five million social interactions~\citep{piao_agentsocietylargescalesimulation_2025}. MF-LLM used mean-field theory and information bottleneck ideas to simulate 300 agents across 20 events and seven fields~\citep{mi_mfllmsimulatingpopulation_2025}. Light Society further proposed a multi-tiered optimization pipeline with prompt caching and knowledge distillation to scale toward one billion agents~\citep{guan_modelingearthscalehumanlike_2025}. These systems reveal a persistent trade-off: very large simulations often sacrifice individual-level fidelity.

\labeledlead{Granularity}{The simulation step determines what kinds of mechanisms remain visible.}
Fine-grained simulations can model turn-by-turn dialogue, such as LLM agents playing trust games~\citep{xie_canlargelanguage_2024}. Coarser simulations can cover longer learning periods; Yuan et al. simulated four students over an entire year using weekly, monthly, and yearly activities~\citep{yuan_simulatinghumanlikelearning_2025}. In education, the appropriate granularity depends on the research question: discourse analysis requires turns, curriculum progression may require days or weeks, and institutional policy may require semesters or years.

Granularity is not simply a computational preference; it changes the ontology of the model. At the turn level, the primary objects are utterances, discourse moves, and immediate feedback. At the lesson level, the primary objects are activities, task sequences, and local learning outcomes. At the semester level, the primary objects are curriculum coverage, accumulated mastery, motivation, and social position. A simulator that collapses all of these into the same step risks either excessive cost or excessive abstraction. AgentSchool therefore supports mixed granularity: an important lesson can be simulated turn by turn, while routine practice can be compressed into state updates with summary evidence.

\labeledlead{Duration}{Educational simulation must cover both moments and trajectories.}
Current educational simulations often last for one lesson, several lessons, or several months. Course-level simulations preserve dialogue reliability by generating every turn. Month-level simulations compress time into hours, days, or weeks to reduce redundant detail while retaining developmental structure.

Duration is essential because many educational effects are delayed. Scaffolding may temporarily increase dependence before enabling independence; repeated low-quality AI feedback may produce fluent but shallow answers; a social exclusion event may later reduce participation in collaborative tasks. These effects cannot be evaluated by inspecting a single generated conversation. AgentSchool records longitudinal histories so that researchers can ask whether short-term interaction patterns accumulate into stable cognitive, social, or institutional outcomes.

\subsection{Simulator Calibration}

\guidingquestion{How can we know whether a simulated school is producing meaningful evidence rather than plausible fiction?}

When simulation is used as an educational tool, evaluation often asks whether the tool supports the learning activity~\citep{emond_cognitivesimulationsadaptive_2023}. When simulation is used for exploration and hypothesis testing, the standard is stricter: outputs must align with the real system at the level relevant to the claim~\citep{alharbi_agentbasedclassroomenvironment_2021}. Otherwise, conclusions drawn from the simulation are unreliable.

Calibration should therefore be defined relative to an intended claim. If the claim concerns classroom discourse, then turn-taking, question types, feedback moves, and student uptake must be calibrated. If the claim concerns learning growth, then mastery trajectories, error patterns, and transfer performance must be calibrated. If the claim concerns institutional reform, then the simulator must reproduce plausible patterns of resource allocation, legitimacy seeking, professional norms, and policy response. A single global realism score is insufficient because realism is multidimensional. AgentSchool treats calibration as a set of alignment checks between simulated observables and either empirical data or theory-derived constraints.

\begin{enumerate}[label=\textbf{M\arabic*.}, leftmargin=*]
    \item \textbf{Prompt and hyperparameter calibration.} Manual prompt engineering and temperature/top-p adjustment are common~\citep{hicke_medsimaisimulationformative_2025,yuan_simulatinghumanlikelearning_2025}, but they are time-intensive, unstable, and difficult to scale.
    \item \textbf{Architectural calibration.} System-level mechanisms, such as complex agent structures~\citep{rao_multiagentsystemcomprehensive_2025} and output validation~\citep{xu_classroomsimulacrabuilding_2025}, can improve stability but remain limited by the underlying model.
    \item \textbf{Data-driven calibration.} Retrieval-augmented generation, supervised fine-tuning, and reinforcement learning can shift model behavior more fundamentally, but require high-quality educational data and substantial technical effort.
\end{enumerate}

Evaluation likewise falls into two complementary families. \textbf{Real-world data matching} measures whether generated outcomes or trajectories fit empirical data~\citep{mi_mfllmsimulatingpopulation_2025}. \textbf{Theory-driven analysis} evaluates whether behavior is consistent with a theoretical framework, often through qualitative or quantitative discourse analysis~\citep{yuan_simulatinghumanlikelearning_2025}. These approaches are not mutually exclusive; alignment can be calculated at multiple levels within a theoretical framework to improve credibility~\citep{xie_canlargelanguage_2024}.

For a set of observables $\mathcal{Z}$, empirical calibration can be expressed as minimizing a distance between simulated and real distributions:
\[
    \min_{\theta} \sum_{z \in \mathcal{Z}} w_z \, D\!\left(P_{\mathrm{sim}}(z \mid \theta), P_{\mathrm{real}}(z)\right),
\]
where $\theta$ denotes simulation parameters, $w_z$ denotes the importance of observable $z$, and $D(\cdot,\cdot)$ may be a distributional distance, a prediction error, or a task-specific divergence. Theory-driven calibration instead constrains acceptable trajectories:
\[
    \mathcal{T}_{\mathrm{valid}}=\left\{ X_{0:H} \; \middle| \; \phi_k(X_{0:H}) \leq \epsilon_k,\; k=1,\ldots,K \right\},
\]
where each $\phi_k$ operationalizes a theoretical constraint, such as gradual mastery growth, ZPD-consistent task difficulty, or plausible social-network evolution. In practice, AgentSchool combines these two approaches: empirical data are used when available, while theory-derived constraints prevent the simulator from treating surface plausibility as sufficient evidence.

\remarkbox{The calibration bottleneck in education is not only technical. Longitudinal educational data are difficult to collect because they involve minors, ethical approval, privacy protection, and long time spans. AgentSchool therefore combines theory-driven structure, controlled scenario validation, and data-grounded calibration where available.}

\section{AgentSchool System}
\label{sec:system}
\sectionsubtitle{AgentSchool is organized as a modular simulation stack. Student agents model cognitive growth, teacher agents model adaptive scaffolding, the scenery generator defines the learning field, and the simulator architecture coordinates interaction across time, scale, and granularity.}

\begin{figure}[h]
    \centering
    \includegraphics[width=\textwidth]{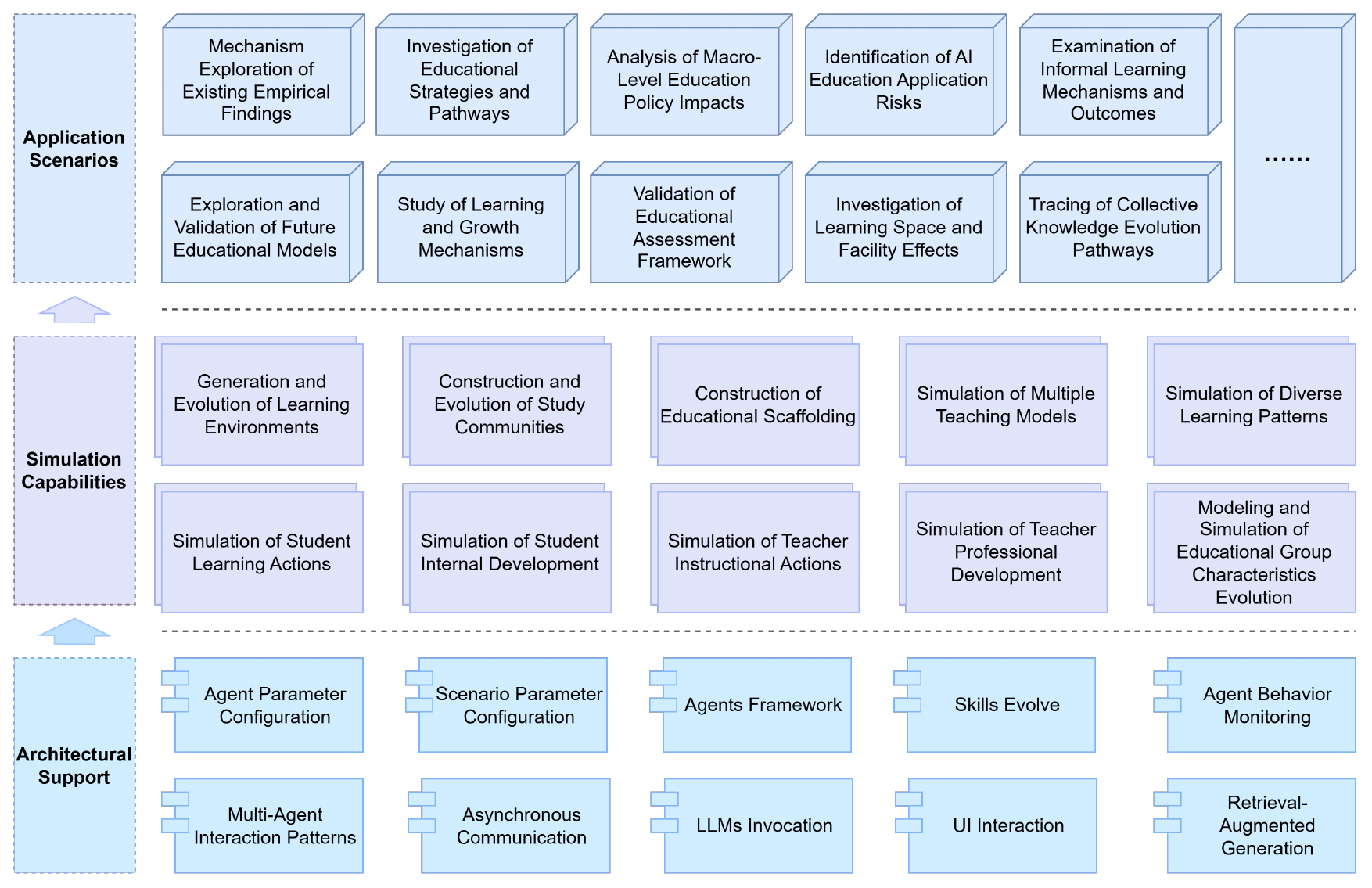}
    \caption{Software Architecture of AgentSchool Platform}
    \label{fig:software_architecture}
\end{figure}

To address the gaps above, AgentSchool proposes a dynamic, cognitively plausible multi-agent educational simulation system through data-driven and LLM-empowered approaches. The system includes cognitively growable student agents, adaptive teacher agents, a configurable scenery generator with planned evolutionary extensions, a high-performance simulator architecture, agent configuration, model configuration, and user interaction modules. The software architecture is illustrated in Figure~\ref{fig:software_architecture}.

The design principle of AgentSchool is to separate \textit{what changes} from \textit{how change is generated}. Student cognition, teacher expertise, scenario conditions, and social relations are represented as explicit state variables. LLMs are then used to generate context-sensitive actions, interpretations, reflections, and discourse under these constraints. This avoids two common extremes. A purely rule-based simulator is often transparent but too rigid to capture the linguistic and social richness of education. A purely prompt-based role-play simulator is expressive but difficult to inspect, calibrate, or reproduce. AgentSchool combines structured state with generative action: the state provides educational semantics, while the LLM provides flexible enactment.

At a high level, one simulation cycle contains four phases. First, the simulator constructs each agent's local observation from the global state and current scenery. Second, agents deliberate and produce actions, such as explanation, question asking, peer response, topic switching, or reflection. Third, the simulator validates and applies these actions to update memories, knowledge graphs, social relations, and scene variables. Fourth, the system logs both the observable event and the hidden state transition, creating an audit trail for later evaluation. This loop can be written as
\[
    o_{i,t} = \Omega_i(X_t, C_t), \quad
    a_{i,t} \sim \pi_i(a \mid o_{i,t}, m_{i,t}; \lambda_i), \quad
    X_{t+1}=F_{\theta}(X_t, \mathbf{a}_t, C_t),
\]
where $\Omega_i$ is the observation function, $\pi_i$ is the agent policy implemented through structured prompts and memory retrieval, $m_{i,t}$ is the agent's accessible memory, $\lambda_i$ encodes role-specific parameters, and $F_{\theta}$ is the simulator transition function.


\begin{figure}[!htb]
\centering
\includegraphics[width=\textwidth]{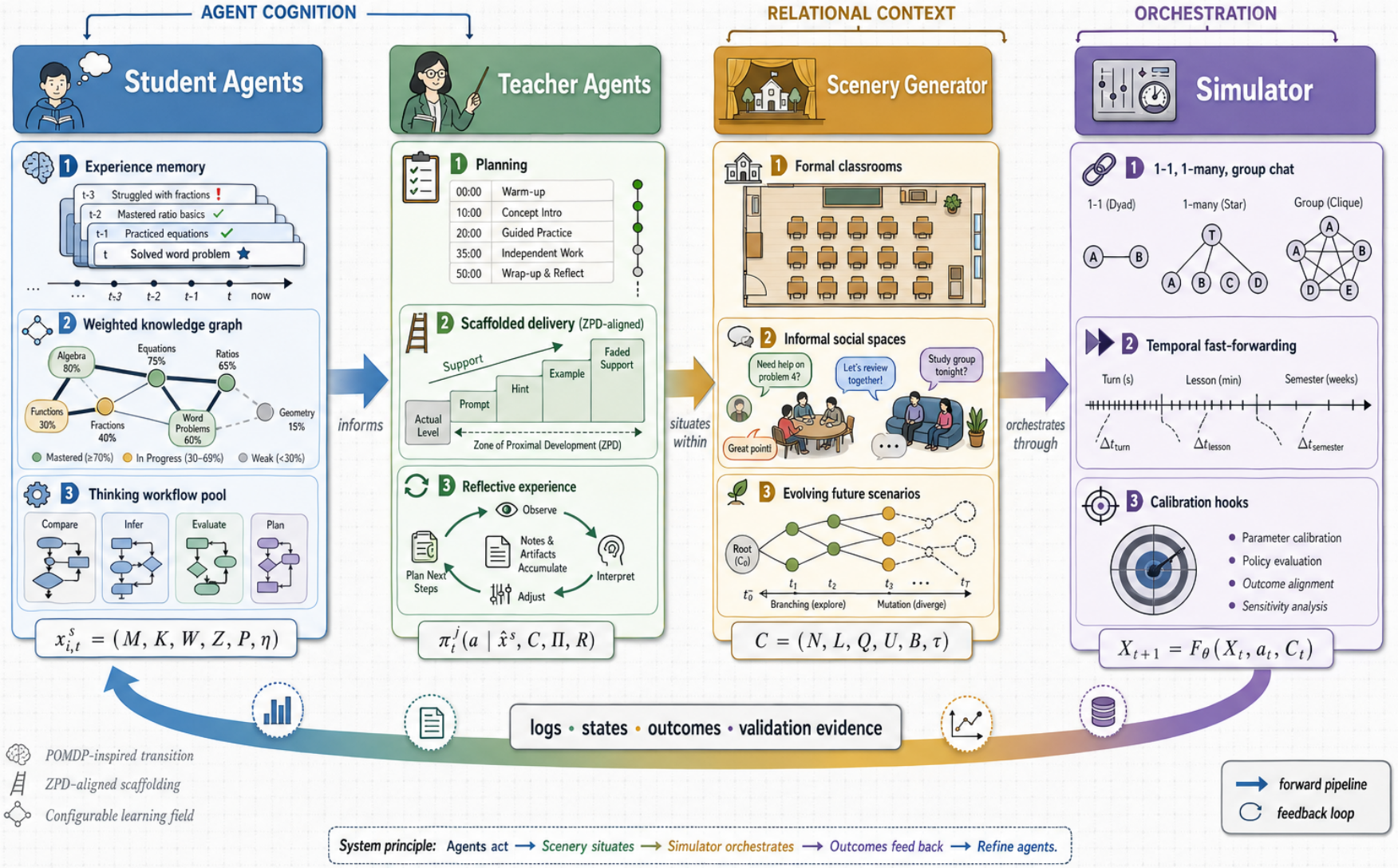}
\caption{Conceptual roadmap of AgentSchool. The platform couples agent cognition, pedagogical action, scenario construction, and simulation orchestration into one inspectable loop.}
\label{fig:agentschool_roadmap}
\end{figure}

\subsection{Student Agents}

\chapterepigraph{The educational process has no end beyond itself; it is its own end.}{John Dewey}{Education as Growth~\citep{dewey2020education}}

AgentSchool treats learning as continuous transformation rather than static knowledge accumulation. Student agents therefore need to model not only what a learner knows, but how knowledge, reasoning habits, and misconceptions reorganize through interaction. Within competency-oriented education, the student state usually includes knowledge and thinking ability~\citep{oecd_21stcenturyskills_2009,ieeesa_competenciesrequestinterest_2015,europeancommission_proposalcouncilrecommendation_2018}. This state has group-level regularities but individual-level variation.

\labeledlead{Internal state}{A student agent maintains prior experience, acquired knowledge, and mastered thinking abilities.}
AgentSchool implements these components as a dialogue memory repository, a weighted subject knowledge graph, and a weighted thinking workflow pool. Stage-specific templates provide group-level regularities. Individual agents are then instantiated through parameterized sampling, producing heterogeneous knowledge weights, learning modalities, and reasoning workflows.

More specifically, the state of student $i$ at time $t$ is represented as
\[
    x^s_{i,t} = \left(M_{i,t}, K_{i,t}, W_{i,t}, Z_{i,t}, P_i, \eta_i \right).
\]
Here $M_{i,t}$ is the episodic memory repository; $K_{i,t}=(V_{i,t},E^K_{i,t},\mu_{i,t})$ is a weighted knowledge graph whose nodes are concepts, edges are prerequisite or semantic relations, and $\mu_{i,t}:V_{i,t}\rightarrow[0,1]$ maps each concept to a mastery score; $W_{i,t}$ is a pool of thinking workflows such as comparison, causal explanation, evidence evaluation, and spatial reasoning; $Z_{i,t}$ stores explicit misconceptions and unstable alternative conceptions; $P_i$ encodes relatively stable learner attributes such as preference, modality, and personality; and $\eta_i$ controls learning sensitivity, forgetting, and response style.

This representation allows AgentSchool to avoid a common weakness of student role-play: the learner is not defined only by a biography prompt. A biography may say that a student is ``weak in geography,'' but it does not specify which concepts are weak, how they are connected, or how a misconception should influence later reasoning. In AgentSchool, weakness can be localized to a node or subgraph; a misconception can be stored as a competing edge or false proposition; a reasoning workflow can succeed in one context and fail in another. The student agent can therefore produce answers that are not merely stylistically student-like, but structurally consistent with its internal cognitive state.

\labeledlead{Growth process}{Learning is modeled as state transition through practice.}
When a student agent receives classroom content, peer communication, or other environmental input, it selects and processes the input through learning modalities. The result updates its memory and knowledge graph. Common learner error types are explicitly modeled to introduce uncertainty and to represent misconceptions or alternative conceptions~\citep{nussbaum_alternativeframeworksconceptual_1982}.

For a knowledge node $v$, mastery update is modeled as a bounded transition:
\[
    \mu_{i,t+1}(v)=\mathrm{clip}_{[0,1]}\left(\mu_{i,t}(v)+\alpha_i q_t(v) r_{i,t}(v)-\beta_i d_t(v)+\gamma_i s_{i,t}(v)\right).
\]
The term $q_t(v)$ represents the quality and relevance of instructional exposure, $r_{i,t}(v)$ represents learner uptake inferred from the student's response, $d_t(v)$ represents decay or interference, and $s_{i,t}(v)$ represents scaffolded support from teachers or peers. The parameters $\alpha_i$, $\beta_i$, and $\gamma_i$ allow heterogeneous growth rates. This formula is not intended to reduce learning to a single number; rather, it provides an explicit update interface so that LLM-generated events can be converted into inspectable state changes.

Misconceptions are represented as structured objects rather than as generic error labels:
\[
    z = \left(v, \hat{p}, p^\ast, c, \rho \right),
\]
where $v$ is the relevant concept, $\hat{p}$ is the learner's current belief, $p^\ast$ is the target scientific or curricular proposition, $c$ is contextual evidence for the misconception, and $\rho\in[0,1]$ is its persistence. This makes it possible to distinguish a random wrong answer from a stable alternative conception. During instruction, the teacher agent can target high-persistence misconceptions through conflict, analogy, demonstration, or guided questioning.

\labeledlead{External expression}{Internal cognitive states are inferred from observable performance.}
Because learning states are internal, continuous, and qualitative, they cannot be directly read from the outside. AgentSchool therefore evaluates student growth through performance in complex contexts, including question answering, discussion, and situated tasks~\citep{annelinAssessmentKeySustainability2022}.

Observable performance is treated as evidence, not as the state itself. Let $y_{i,t}$ denote a student utterance or task output. The simulator estimates an evidence function
\[
    \hat{x}^s_{i,t} = \Psi(y_{i,0:t}, C_{0:t}, \mathcal{B}^{\mathrm{eval}}),
\]
where $\Psi$ maps accumulated behavior, scenario history, and evaluation rubric $\mathcal{B}^{\mathrm{eval}}$ to an inferred learner profile. The gap between the internal state $x^s_{i,t}$ and inferred state $\hat{x}^s_{i,t}$ is useful: a large gap may indicate hidden knowledge, performance anxiety, overconfident guessing, or insufficient assessment evidence. This distinction is pedagogically important because real teachers also infer understanding from incomplete observations.

\remarkbox{The key design choice is to make student agents \textbf{growable}. They are not fixed personas attached to prompts; they are stateful learners whose memories, knowledge structures, reasoning workflows, and misconceptions change across simulated educational experience.}

\subsection{Teacher Agents}


\guidingquestion{Can a teacher agent model pedagogy as adaptive support rather than scripted delivery?}

The design of teacher agents follows student-centered pedagogy. Dewey's ``learning by doing'' positions teachers as facilitators of growth. Vygotsky's Zone of Proximal Development (ZPD) identifies the space between what learners can do independently and what they can achieve with assistance~\citep{vygotskyumstvennoie1935}. Wood, Bruner, and Ross's scaffolding theory describes how temporary support can help learners move through this zone toward independence~\citep{wood_roletutoringproblem_1976}.

AgentSchool teacher agents execute a complete instructional cycle:
\begin{enumerate}[label=\textbf{Step \arabic*.}, leftmargin=*]
    \item \textbf{Planning.} The planning module generates learning objectives, instructional activities, and time allocation.
    \item \textbf{Scaffolded delivery.} The agent iteratively performs five instructional moves: establishing scaffolds, contextual engagement, independent exploration, collaborative learning, and outcome assessment.
    \item \textbf{Adaptive exploration.} During instruction, the agent combines teaching methods and scaffold types to explore alternative pedagogical pathways.
    \item \textbf{Reflective growth.} The agent records dialogue, student feedback, and performance data, then updates its experiential knowledge base.
\end{enumerate}

Each teacher agent maintains two memory systems. A declarative knowledge base stores subject content and pedagogical content knowledge derived from instructional materials. An experiential knowledge base accumulates practical teaching knowledge from simulated lessons. Parameterized pedagogical principles shape how the agent interprets its role, manages classroom interaction, and adapts to student states.

The teacher state is represented as
\[
    x^t_{j,t}=\left(D_{j,t}, R_{j,t}, B_j, \Pi_{j,t}, \xi_j\right),
\]
where $D_{j,t}$ is declarative subject and pedagogical knowledge, $R_{j,t}$ is reflective experience accumulated from previous simulations, $B_j$ is the teacher's pedagogical belief profile, $\Pi_{j,t}$ is the current lesson plan, and $\xi_j$ contains style and constraint parameters such as tolerance for uncertainty, preference for direct instruction, or emphasis on collaboration. Given observations of student states, the teacher selects an instructional action
\[
    a^t_{j,t} \sim \pi^t_j\!\left(a \mid \hat{x}^s_{1:n,t}, C_t, \Pi_{j,t}, R_{j,t}\right).
\]
Actions include explanation, questioning, demonstration, grouping, hinting, feedback, affective encouragement, misconception challenge, and task redesign.

ZPD alignment is operationalized by comparing task difficulty with estimated learner readiness. Let $d(a,v)$ denote the cognitive demand of instructional action $a$ on concept $v$, and let $\mu_{i,t}(v)$ denote student $i$'s current mastery. A simple ZPD compatibility score can be written as
\[
    \mathrm{ZPD}_{i,t}(a,v)=\exp\!\left(-\frac{\left(d(a,v)-(\mu_{i,t}(v)+\delta_i)\right)^2}{2\sigma_i^2}\right),
\]
where $\delta_i$ approximates the additional capacity available under assistance and $\sigma_i$ controls the acceptable challenge band. The teacher does not need to maximize this expression mechanically, but the score provides an evaluable definition of whether the teacher is selecting tasks that are too easy, too difficult, or appropriately scaffolded.

Reflection updates the experiential knowledge base after a lesson. The teacher agent summarizes what was attempted, what evidence was observed, which misconceptions persisted, and which scaffolds appeared effective. The update can be represented as
\[
    R_{j,t+1}=R_{j,t}\cup \left\{(\Pi_{j,t}, C_t, \hat{x}^s_{1:n,t}, \mathbf{a}^t_{j,0:t}, \Delta K_{1:n,t}, \ell_t)\right\},
\]
where $\Delta K_{1:n,t}$ denotes changes in student knowledge graphs and $\ell_t$ denotes lesson-level evaluation. This design makes teacher growth endogenous: a teacher agent can become more effective because prior simulation experience changes future planning and scaffolding choices.

\begin{remarkboxenv}{Teacher Agent Design Principle}
The teacher is not a prompt wrapper around a lesson plan. It is an evolving pedagogical agent that plans, teaches, observes, reflects, and revises its future practice.
\end{remarkboxenv}

\subsection{Scenery Generator}



AgentSchool defines scenery as the field in which learning occurs. A scenery is a meaning-laden relational network constituted by sociocultural context, physical environment, interpersonal relationship, tools and symbols, and practical activities. Educational participants are shaped by this field, but they also reshape it through participation.

In implementation, scenery is not a decorative background. It is a parameterized object that constrains observation, action, and interpretation. A scenery instance contains spatial structure, participant roles, resource availability, communication channels, temporal schedule, participation norms, and evaluation expectations. For example, the same geography topic can be enacted as a lecture, a map-reading group task, a debate about regional development, an online asynchronous discussion, or an informal peer explanation during recess. Each setting changes who can speak, what evidence is available, whether peer support is likely, and how the teacher can intervene.

\labeledlead{Predefined scenery}{The platform provides structured templates distilled from existing educational settings.}
Users can instantiate traditional classrooms, open discussion classrooms, online learning spaces, and other settings by adjusting parameters. In a lecture-based classroom, student distribution is usually uniform, language flow is mostly one-way, and interaction is radial. In an open small-group classroom, spatial distribution is less uniform, interaction is denser, and communities of practice are easier to form.

Each predefined template specifies a default interaction graph $G^C=(V^C,E^C)$. In a traditional classroom, the graph is teacher-centered: most high-probability edges connect the teacher to individual students, while peer-to-peer edges are sparse. In an open classroom, the graph is clustered: students have dense local edges within groups and weaker edges across groups. In an online setting, the graph may be asynchronous and mediated by artifacts. These graph priors are not fixed outcomes; they initialize probabilities that can change as agents interact. A quiet student may become central after repeatedly helping peers, while an aggressive student may lose social ties even in a densely connected layout.

\labeledlead{Informal scenery}{AgentSchool treats social chat as part of education.}
Recess conversations, peer gossip, hobby groups, and spontaneous topic discussions are informal learning spaces where knowledge and norms flow implicitly. Many educational simulators ignore these scenes because they are not formal instruction, but they are central to belonging, identity, social norms, and classroom governance.

Informal scenery is modeled with weaker institutional scripts and stronger agent autonomy. Instead of a teacher-defined lesson plan, topics emerge from interests, personality, recent events, and social ties. The simulator tracks topic formation, joining, withdrawal, emotional tone, and relationship change. This matters because informal social structure can later shape formal learning. A student who is socially isolated may avoid group discussion; a student who becomes an opinion leader may influence attitudes toward an AI tool or a teacher's new pedagogy. Treating informal scenes as part of the educational system makes AgentSchool more suitable for studying classroom climate and school governance.

\labeledlead{Evolutionary scenery}{Future education may require scenarios that do not yet exist.}
We plan to evolve candidate educational settings through an Alpha-Evolve-like process. An LLM ensemble constructs or modifies scenario candidates. An evaluator pool scores each candidate using automated criteria and progressively harder tests. A scenery database stores validated settings, which can then inspire subsequent generations.

The evolutionary scenery pipeline is intended to support counterfactual design. A candidate scenery $C'$ is generated by mutating an existing scenery $C$ along dimensions such as spatial layout, teacher role, AI-tool access, grouping policy, or assessment mode. It is then evaluated by a vector of criteria
\[
    Q(C') = \left(q_{\mathrm{learning}}, q_{\mathrm{equity}}, q_{\mathrm{feasibility}}, q_{\mathrm{safety}}, q_{\mathrm{novelty}}\right).
\]
Sceneries that improve one dimension while severely harming another can be filtered or sent to human review. This procedure does not claim to automatically discover the future of education; rather, it provides a disciplined way to generate and stress-test possibilities that would be costly or unethical to try first in real schools.

\remarkbox{Scenery also functions as a validation instrument. By examining how student agents perform and interact within a specific learning field, researchers can infer displayed cognitive states and compare them with internal agent states. This enables theory-driven validation even when real-world data are limited.}

\subsection{Simulator Architecture}

AgentSchool's simulator is tailored to educational dynamics: it must support fine-grained interaction, long-horizon growth, and efficient orchestration across many agents.

\labeledlead{State transition}{Student development is modeled as an interaction-driven, partially observable state-transition process.}
Students transition between states through actions that involve the external environment. These actions include formal instruction, peer discussion, teacher-student after-class exchange, and other forms of learning-field interaction. The simulator also exposes evaluation signals that can later support agentic reinforcement-learning extensions, but the present paper uses the formalism primarily to make state updates inspectable.

At the individual level, a student can be formulated as a POMDP-inspired process
\[
    \mathcal{M}_i=\left(\mathcal{X}_i,\mathcal{U}_i,\mathcal{P}_i,\mathcal{J}_i,\Omega_i\right),
\]
where $\mathcal{X}_i$ is the student state space, $\mathcal{U}_i$ is the action space, $\mathcal{P}_i$ is the transition kernel induced by learning and forgetting, $\mathcal{J}_i$ is a task-dependent evaluation signal, and $\Omega_i$ is the observation function. At the classroom level, these processes are coupled because one student's action can become another student's observation. A peer explanation, for instance, updates the speaker's memory through articulation and the listener's knowledge graph through reception. This coupling is what makes multi-agent educational simulation different from independent tutoring simulation.

\labeledlead{Interaction patterns}{Classroom interaction can switch as the learning field evolves.}
AgentSchool abstracts classroom communication into three core patterns: one-to-one interaction, one-to-many broadcasting, and unstructured group chat. Lecture, questioning, group discussion, peer learning, and after-class conversation can be composed through these primitives.

These primitives are implemented as communication operators. A one-to-many broadcast applies the same teacher utterance to multiple student observations but allows heterogeneous interpretation because each student has different prior knowledge and attention. A one-to-one exchange increases conversational depth and diagnostic precision, but is expensive at scale. Group chat creates a shared discourse history and allows emergent peer influence, but also introduces topic drift and unequal participation. By composing these operators, AgentSchool can model conventional lecture, Socratic questioning, group inquiry, peer tutoring, and informal discussion without creating a separate simulator for each pedagogy.

\labeledlead{Learning communities}{The simulator can construct task-oriented groups with shared goals and collective identity.}
Within these communities, members share practice, knowledge, and expertise. They gradually move from peripheral participation toward core participation, allowing viable pedagogical models to emerge from repeated interaction.

Community membership is represented as a dynamic relation rather than a static group label. Let $g_{ij,t}$ denote the tie strength between students $i$ and $j$. After an interaction event $e_t$, the tie can be updated by
\[
    g_{ij,t+1}=\mathrm{clip}_{[0,1]}\left(g_{ij,t}+u(e_t)-v(e_t)+h_{ij,t}\right),
\]
where $u(e_t)$ captures affiliative or productive interaction, $v(e_t)$ captures conflict or exclusion, and $h_{ij,t}$ captures homophily or shared history. Learning communities emerge when dense, stable ties coincide with shared goals and repeated knowledge exchange. This enables the simulator to track whether a collaborative learning design actually creates a community of practice or merely places students into nominal groups.

\labeledlead{Efficiency}{Computational optimizations reduce runtime while preserving the level of detail needed for validation.}
AgentSchool supports temporal granularities ranging from turn-by-turn classroom dialogue to semester-long progression, and population scales ranging from individual case studies to institutional populations.

Efficiency is achieved by adaptive resolution. The simulator can allocate more LLM calls to educationally consequential moments, such as misconception diagnosis, teacher reflection, conflict resolution, or high-stakes assessment, while compressing routine transitions into structured updates. It also supports caching of stable agent profiles, retrieval of relevant memory rather than full-history prompting, and event-level logging instead of repeated natural-language summaries. These optimizations are necessary because educational simulation must often run multiple counterfactual conditions rather than a single narrative trace.

\begin{remarkboxenv}{System Summary}
AgentSchool combines \textbf{cognitive growth} (student agents), \textbf{adaptive scaffolding} (teacher agents), \textbf{relational context} (scenery), and \textbf{multi-scale orchestration} (simulator architecture). This combination is what allows the platform to serve as a wind tunnel rather than a simple role-play environment.
\end{remarkboxenv}


\section{Preliminary Result}
\label{sec:result}
\sectionsubtitle{We conduct preliminary validation in two complementary settings: formal lesson simulation and informal recess social simulation. The goal is to examine whether AgentSchool can produce interpretable, educationally meaningful dynamics across both instructional and social scenes.}

A critical question for any simulator is whether its generated data align with the real world. This alignment can be understood at two levels. \textbf{Phenomenological alignment} asks whether the simulator reproduces recognizable patterns and mechanisms. \textbf{Statistical alignment} asks whether the simulator quantitatively matches real-world distributions. Most existing LLM-based simulations can only claim the former. Our preliminary experiments therefore focus on whether AgentSchool generates interpretable and theoretically grounded phenomena in formal and informal educational contexts.

We emphasize that these experiments are preliminary validation studies rather than final claims of ecological equivalence. Their purpose is to test whether the proposed architecture produces richer and more diagnosable educational dynamics than baseline role-play approaches under controlled conditions. A useful simulator should satisfy at least three requirements at this stage. First, it should produce observable differences when internal mechanisms differ; otherwise the architecture is not doing explanatory work. Second, its outputs should be interpretable through educational theory rather than only through generic language quality. Third, it should reveal both positive and negative outcomes, including misconceptions, low-mastery regions, and social exclusion, because a simulator that only generates successful learning is not useful for risk analysis.

Accordingly, we evaluate AgentSchool in two settings. The lesson simulation examines formal instruction, where the key question is whether student and teacher agents generate meaningful cognitive trajectories under controlled pedagogy. The social simulation examines informal classroom life, where the key question is whether peer interaction produces plausible network evolution. Together, these settings cover two major sources of educational change: explicit teaching and implicit social participation.

\paragraph{Evaluation protocol and claim boundary.}
The present experiments should be read as \textit{within-simulator diagnostic validation}. Dialogue traces, event logs, and state updates are converted into knowledge nodes, mastery estimates, and misconception objects through a fixed analysis procedure. These indicators are useful for comparing simulation conditions because the same procedure is applied across models and pairings, but they are not yet calibrated against longitudinal classroom data. We therefore avoid claiming ecological equivalence. The evidence supports a narrower claim: structured stateful agents produce more inspectable and educationally interpretable traces than prompt-only role play under the same simulated task.

\begin{figure}[!htb]
\centering
\includegraphics[width=\textwidth]{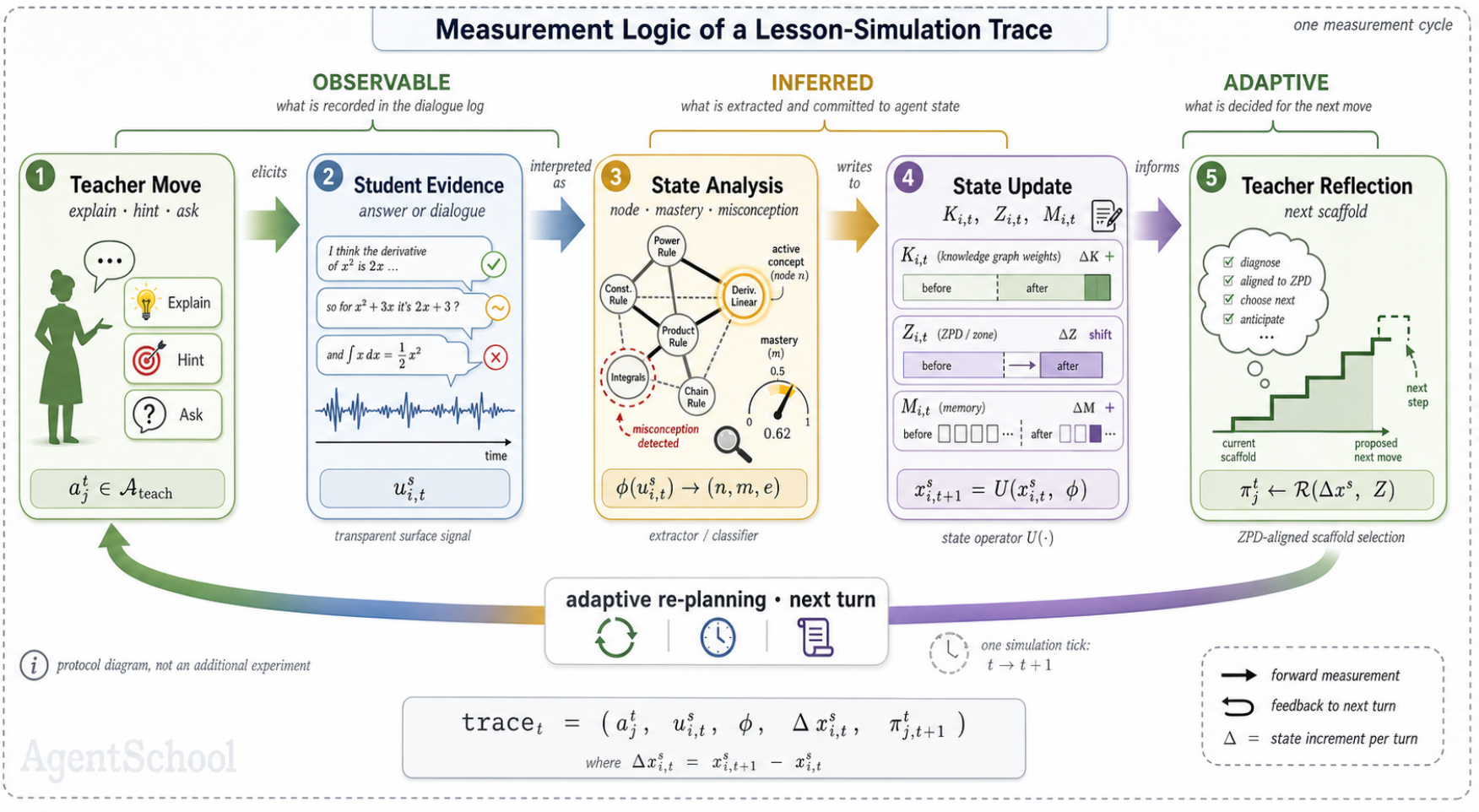}
\caption{Measurement logic of a lesson-simulation trace. The figure illustrates how observable dialogue is converted into inspectable state changes and subsequent teacher adaptation; it is a protocol diagram rather than an additional experiment.}
\label{fig:lesson_trace_protocol}
\end{figure}

\subsection{Lesson Simulation}
\guidingquestion{Can AgentSchool represent learning gains, misconceptions, and teacher effects more richly than baseline role-play simulators?}

We simulated a lecture-based middle school geography classroom. With standardized initial knowledge points, we conducted a $2 \times 3$ controlled experiment: student simulator ($S_{\text{ours}}$ vs. $S_{\text{base}}$) crossed with teacher type ($T_{\text{ours}}$ vs. $T_{\text{base}}$ vs. $T_{\text{scripted}}$)~\citep{jiang_evolutionaryreinforcementlearning_2025}. Here ``base'' denotes the prior role-play-style simulator and teacher baseline used for comparison, while $T_{\text{scripted}}$ denotes a fixed lesson script. Classroom dialogues were used as evidence for constructing and analyzing knowledge graphs under the ZPD perspective.

The experiment is designed to isolate two architectural factors. The student factor asks whether a cognitively growable student agent produces different learning traces from a baseline student simulator. The teacher factor asks whether an adaptive, reflective teacher agent produces different outcomes from a baseline teacher model and a scripted teacher. The scripted teacher is included because a fixed high-quality lesson may perform well on content coverage but may fail to respond to learner state. The crossed design allows us to examine not only the main effect of each component, but also the interaction between student representation and teacher adaptation.

Let $\mathcal{V}_{m,c}$ be the set of knowledge nodes generated for model $m$ under condition $c$, and let $\mu(v)$ be the final mastery score of node $v$. The four reported indicators are defined as
\[
    \mathrm{AvgMastery}_{m,c}=\frac{1}{|\mathcal{V}_{m,c}|}\sum_{v\in\mathcal{V}_{m,c}}\mu(v),
\]
\[
    \mathrm{HighNodes}_{m,c}=\sum_{v\in\mathcal{V}_{m,c}}\mathbb{I}[\mu(v)\geq0.8], \qquad
    \mathrm{LowNodes}_{m,c}=\sum_{v\in\mathcal{V}_{m,c}}\mathbb{I}[\mu(v)<0.2],
\]
and
\[
    \mathrm{Misc}_{m,c}=|\mathcal{Z}^{\mathrm{mis}}_{m,c}|,
\]
where $\mathcal{Z}^{\mathrm{mis}}_{m,c}$ denotes the set of detected misconception objects. These metrics intentionally combine achievement and diagnostic richness. Average mastery measures overall learning level, high nodes measure confident acquisition, low nodes reveal unresolved weak areas, and misconceptions reveal structured alternative understanding rather than simple failure.

Because different backbone models can generate knowledge graphs of different sizes, raw node counts should be interpreted as diagnostic counts rather than normalized achievement rates. The current experiment does not report total node counts uniformly across all conditions; we therefore use High Nodes and Low Nodes to inspect trace richness and unresolved weakness, while relying on Avg. Mastery for the more directly comparable scalar outcome.

We compare teaching effectiveness across five backbone LLMs using four indicators:
\begin{itemize}
    \item \textbf{Avg. Mastery}: mean mastery score across generated knowledge nodes.
    \item \textbf{High Nodes}: number of nodes with mastery $\geq 0.8$.
    \item \textbf{Low Nodes}: number of nodes with mastery $<0.2$.
    \item \textbf{Misc.}: number of misconceptions detected in the learner state.
\end{itemize}


\begin{table}[htb!]
\centering
\caption{Comparison of lesson-simulation diagnostics across student and teacher pairings.}
\label{tab:teacher_comparison}
\begingroup
\footnotesize
\setlength{\tabcolsep}{0pt}
\renewcommand{\arraystretch}{1.12}
\def\sp{0.55mm}
\def\metricboxw{12.8mm}
\newcommand{\metricbox}[4]{\tikz[baseline=(metric.base)]{\node[
    rounded corners=1.8pt,
    inner xsep=1.5pt,
    inner ysep=0.7pt,
    minimum width=\metricboxw,
    minimum height=3.35mm,
    outer ysep=0.25pt,
    fill={rgb,255:red,#1;green,#2;blue,#3},
    text=black
] (metric) {\strut #4};}}
\newcommand{\modelblock}[1]{\multirow{6}{*}[-1mm]{\parbox[m][][c]{2.35cm}{\centering\textbf{#1}}}}
\begin{adjustbox}{scale=1.0,center}
\begin{tabular}{@{}>{\centering\arraybackslash}p{2.35cm}@{\hspace{4pt}}>{\centering\arraybackslash}p{1.55cm}@{\hspace{4pt}}>{\centering\arraybackslash}p{1.85cm}@{\hspace{5pt}}c@{\hspace{4pt}}c@{\hspace{4pt}}c@{\hspace{4pt}}c@{}}
\toprule
\textbf{Model} & \textbf{Student} & \textbf{Teacher} & \makecell{\textbf{Avg.}\\\textbf{Mastery}} & \makecell{\textbf{High}\\\textbf{Nodes}} & \makecell{\textbf{Low}\\\textbf{Nodes}} & \textbf{Misc.} \\

\midrule
\modelblock{Claude-sonnet-4}
& $S_{\text{ours}}$ & $T_{\text{ours}}$ & \metricbox{159}{206}{174}{0.31} & \metricbox{244}{217}{214}{6} & \metricbox{232}{113}{105}{30} & \metricbox{240}{177}{172}{12} \\
& $S_{\text{ours}}$ & $T_{\text{base}}$ & \metricbox{184}{218}{194}{0.29} & \metricbox{247}{238}{235}{7} & \metricbox{237}{150}{145}{41} & \metricbox{239}{170}{165}{11} \\
& $S_{\text{ours}}$ & $T_{\text{scripted}}$ & \metricbox{82}{170}{112}{\textbf{0.37}} & \metricbox{193}{222}{201}{\textbf{10}} & \metricbox{230}{92}{84}{24} & \metricbox{240}{177}{172}{12} \\
& $S_{\text{base}}$ & $T_{\text{ours}}$ & \metricbox{233}{116}{109}{0.13} & \metricbox{230}{92}{84}{0} & \metricbox{244}{215}{212}{60} & \metricbox{232}{106}{99}{2} \\
& $S_{\text{base}}$ & $T_{\text{base}}$ & \metricbox{234}{128}{121}{0.14} & \metricbox{230}{92}{84}{0} & \metricbox{239}{171}{166}{47} & \metricbox{232}{106}{99}{2} \\
& $S_{\text{base}}$ & $T_{\text{scripted}}$ & \metricbox{234}{128}{121}{0.14} & \metricbox{230}{92}{84}{0} & \metricbox{242}{198}{194}{55} & \metricbox{232}{106}{99}{2} \\[\sp]

\midrule
\modelblock{DeepSeek-V3.2}
& $S_{\text{ours}}$ & $T_{\text{ours}}$ & \metricbox{248}{248}{246}{0.24} & \metricbox{237}{243}{237}{8} & \metricbox{245}{219}{216}{61} & \metricbox{230}{92}{84}{0} \\
& $S_{\text{ours}}$ & $T_{\text{base}}$ & \metricbox{133}{194}{153}{\textbf{0.33}} & \metricbox{247}{238}{235}{7} & \metricbox{232}{113}{105}{30} & \metricbox{237}{149}{143}{8} \\
& $S_{\text{ours}}$ & $T_{\text{scripted}}$ & \metricbox{197}{224}{205}{0.28} & \metricbox{148}{201}{166}{\textbf{12}} & \metricbox{243}{209}{205}{58} & \metricbox{242}{198}{194}{15} \\
& $S_{\text{base}}$ & $T_{\text{ours}}$ & \metricbox{230}{92}{84}{0.11} & \metricbox{230}{92}{84}{0} & \metricbox{243}{202}{198}{56} & \metricbox{232}{106}{99}{2} \\
& $S_{\text{base}}$ & $T_{\text{base}}$ & \metricbox{231}{104}{96}{0.12} & \metricbox{230}{92}{84}{0} & \metricbox{238}{161}{155}{44} & \metricbox{232}{106}{99}{2} \\
& $S_{\text{base}}$ & $T_{\text{scripted}}$ & \metricbox{230}{92}{84}{0.11} & \metricbox{230}{92}{84}{0} & \metricbox{241}{188}{184}{52} & \metricbox{232}{106}{99}{2} \\[\sp]

\midrule
\modelblock{Qwen3-32B}
& $S_{\text{ours}}$ & $T_{\text{ours}}$ & \metricbox{184}{218}{194}{\textbf{0.29}} & \metricbox{237}{243}{237}{\textbf{8}} & \metricbox{236}{147}{141}{40} & \metricbox{246}{227}{224}{19} \\
& $S_{\text{ours}}$ & $T_{\text{base}}$ & \metricbox{222}{236}{225}{0.26} & \metricbox{244}{217}{214}{6} & \metricbox{238}{157}{152}{43} & \metricbox{246}{234}{231}{20} \\
& $S_{\text{ours}}$ & $T_{\text{scripted}}$ & \metricbox{197}{224}{205}{0.28} & \metricbox{240}{175}{170}{4} & \metricbox{233}{116}{109}{31} & \metricbox{239}{170}{165}{11} \\
& $S_{\text{base}}$ & $T_{\text{ours}}$ & \metricbox{233}{116}{109}{0.13} & \metricbox{230}{92}{84}{0} & \metricbox{240}{178}{173}{49} & \metricbox{232}{106}{99}{2} \\
& $S_{\text{base}}$ & $T_{\text{base}}$ & \metricbox{238}{164}{159}{0.17} & \metricbox{230}{92}{84}{0} & \metricbox{231}{102}{95}{27} & \metricbox{232}{106}{99}{2} \\
& $S_{\text{base}}$ & $T_{\text{scripted}}$ & \metricbox{238}{164}{159}{0.17} & \metricbox{230}{92}{84}{0} & \metricbox{237}{150}{145}{41} & \metricbox{232}{106}{99}{2} \\[\sp]

\midrule
\modelblock{GPT-5}
& $S_{\text{ours}}$ & $T_{\text{ours}}$ & \metricbox{210}{230}{215}{\textbf{0.27}} & \metricbox{82}{170}{112}{\textbf{15}} & \metricbox{221}{235}{224}{77} & \metricbox{245}{220}{217}{18} \\
& $S_{\text{ours}}$ & $T_{\text{base}}$ & \metricbox{245}{224}{221}{0.22} & \metricbox{193}{222}{201}{10} & \metricbox{144}{199}{162}{98} & \metricbox{195}{223}{203}{29} \\
& $S_{\text{ours}}$ & $T_{\text{scripted}}$ & \metricbox{242}{200}{196}{0.20} & \metricbox{215}{232}{219}{9} & \metricbox{89}{173}{118}{113} & \metricbox{82}{170}{112}{\textbf{44}} \\
& $S_{\text{base}}$ & $T_{\text{ours}}$ & \metricbox{237}{152}{146}{0.16} & \metricbox{237}{154}{149}{3} & \metricbox{82}{170}{112}{\textbf{115}} & \metricbox{232}{106}{99}{2} \\
& $S_{\text{base}}$ & $T_{\text{base}}$ & \metricbox{242}{200}{196}{0.20} & \metricbox{237}{243}{237}{8} & \metricbox{206}{228}{212}{81} & \metricbox{232}{106}{99}{2} \\
& $S_{\text{base}}$ & $T_{\text{scripted}}$ & \metricbox{240}{176}{171}{0.18} & \metricbox{242}{196}{192}{5} & \metricbox{217}{233}{221}{78} & \metricbox{232}{106}{99}{2} \\[\sp]

\midrule
\modelblock{Gemini-2.5-flash}
& $S_{\text{ours}}$ & $T_{\text{ours}}$ & \metricbox{235}{242}{236}{0.25} & \metricbox{247}{238}{235}{7} & \metricbox{242}{198}{194}{55} & \metricbox{239}{170}{165}{11} \\
& $S_{\text{ours}}$ & $T_{\text{base}}$ & \metricbox{248}{248}{246}{0.24} & \metricbox{244}{217}{214}{6} & \metricbox{240}{181}{177}{50} & \metricbox{239}{170}{165}{11} \\
& $S_{\text{ours}}$ & $T_{\text{scripted}}$ & \metricbox{222}{236}{225}{\textbf{0.26}} & \metricbox{215}{232}{219}{\textbf{9}} & \metricbox{237}{154}{148}{42} & \metricbox{232}{106}{99}{2} \\
& $S_{\text{base}}$ & $T_{\text{ours}}$ & \metricbox{237}{152}{146}{0.16} & \metricbox{232}{113}{106}{1} & \metricbox{235}{242}{236}{73} & \metricbox{232}{106}{99}{2} \\
& $S_{\text{base}}$ & $T_{\text{base}}$ & \metricbox{238}{164}{159}{0.17} & \metricbox{230}{92}{84}{0} & \metricbox{245}{219}{216}{61} & \metricbox{232}{106}{99}{2} \\
& $S_{\text{base}}$ & $T_{\text{scripted}}$ & \metricbox{242}{200}{196}{0.20} & \metricbox{240}{175}{170}{4} & \metricbox{240}{178}{173}{49} & \metricbox{232}{106}{99}{2} \\

\bottomrule
\end{tabular}
\end{adjustbox}
\endgroup
\medskip
\parbox{0.78\textwidth}{\footnotesize\raggedright
\textit{Note.} High mastery nodes $\geq 0.8$; Low mastery nodes $< 0.2$; Misc. = Misconceptions; base = prior baseline simulator/teacher. Cell colors are normalized within each metric column: larger values are greener, smaller values are redder, and mid-range values approach white.}
\end{table}

\begin{figure}[!htbp]
    \centering
    \includegraphics[width=\textwidth]{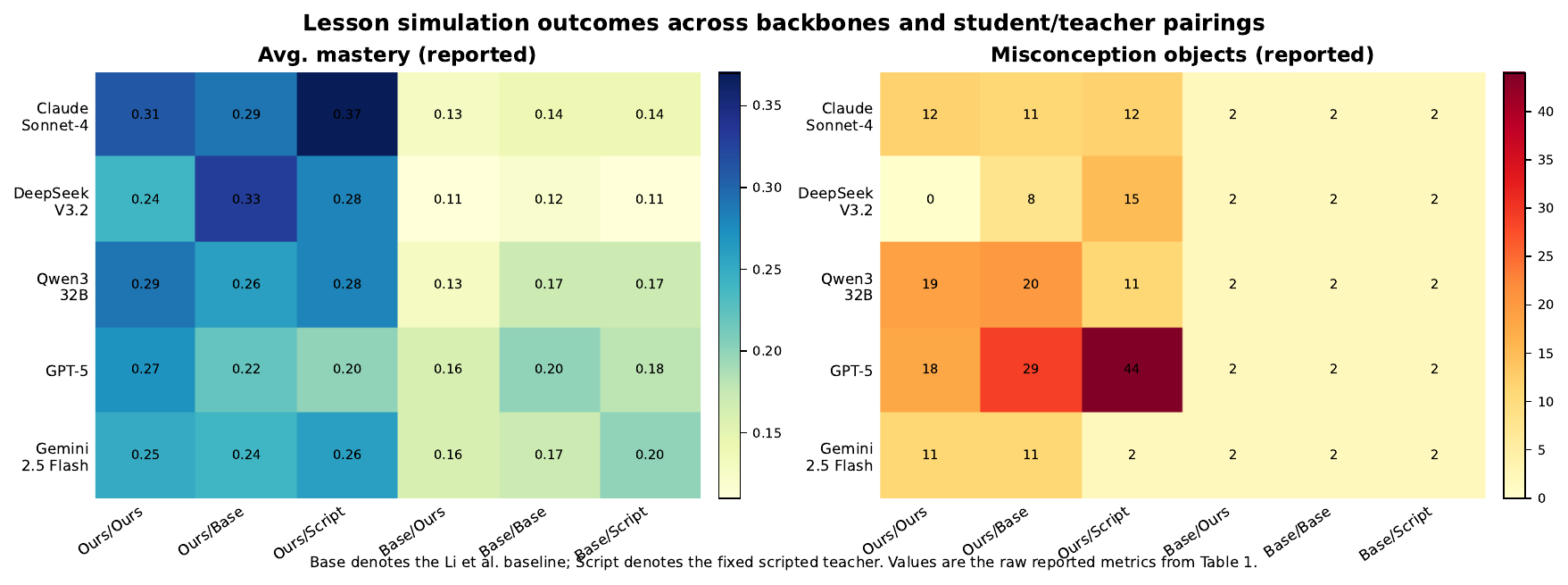}
    \caption{Descriptive heatmaps of the reported lesson-simulation outcomes. The figure visualizes the raw Avg. Mastery and Misc. values from Table~\ref{tab:teacher_comparison}; it is not a normalized performance plot because total generated knowledge-node counts are not uniformly reported.}
    \label{fig:lesson_outcome_heatmaps}
\end{figure}

Table~\ref{tab:teacher_comparison} presents diagnostic lesson outcomes across five backbone models and six student-teacher pairing conditions. Three patterns are especially informative.

\labeledlead{Student simulator effect}{$S_{\text{ours}}$ produces more differentiated learning traces than $S_{\text{base}}$.}
Across backbone models and matched teacher configurations, $S_{\text{ours}}$ achieves higher average mastery scores than $S_{\text{base}}$. It also produces more high-mastery nodes in most conditions. Low-node counts are more mixed because raw counts are affected by the size and granularity of the generated knowledge graph, which is why we treat them as diagnostic evidence rather than normalized failure rates. $S_{\text{base}}$ produces near-zero high-mastery nodes in many conditions, suggesting that it struggles to represent meaningful knowledge acquisition under instructional scaffolding.

Misconception patterns are also revealing. $S_{\text{ours}}$ usually produces a wider range of misconceptions, while $S_{\text{base}}$ often yields the same small count. Although fewer misconceptions may appear better at first glance, such uniformity suggests limited representation of heterogeneous prior knowledge. The richer misconception profiles of $S_{\text{ours}}$ are more consistent with constructivist accounts in which learners naturally form partial or erroneous schemas.

This result should not be interpreted as ``more misconceptions are always better.'' In real education, the desirable outcome is not to maximize misconceptions, but to reveal them accurately when they exist. A simulator that never generates misconceptions may be optimistic rather than faithful. Conversely, a simulator that generates many incoherent misconceptions may be noisy rather than useful. The value of $S_{\text{ours}}$ lies in the combination of higher mastery and richer misconception structure: the agent can both learn and expose partial understanding. This is precisely the kind of trace that allows a teacher agent to diagnose and intervene.

\labeledlead{Teacher agent effect}{Teacher differences are meaningful but backbone-dependent.}
Under $S_{\text{ours}}$, $T_{\text{ours}}$ achieves the highest average mastery on Qwen3-32B and GPT-5, while $T_{\text{base}}$ leads on DeepSeek-V3.2 and $T_{\text{scripted}}$ leads on Claude-sonnet-4 and Gemini-2.5-flash. This pattern is important: the current evidence does not support a blanket claim that the proposed teacher dominates every baseline. Instead, it suggests that adaptive teaching is most informative when the student simulator exposes enough cognitive detail for adaptation to matter, and that a scripted lesson can remain competitive when the content path is stable.

Because this table reports outcome-level diagnostics rather than action-level ZPD scores, we interpret teacher adaptation conservatively. The ZPD claim is supported by the architecture and by outcome patterns that are consistent with adaptive scaffolding, but stronger validation will require logging task difficulty, estimated learner readiness, and teacher response after each detected misconception.

The comparison with $T_{\text{scripted}}$ is particularly important. A scripted lesson can appear strong if evaluated only by content coverage or surface coherence. However, education is not only delivery; it is contingent response. When a student reveals a misconception, the teacher must decide whether to explain again, ask a diagnostic question, provide an analogy, adjust task difficulty, or invite peer explanation. A script cannot reliably make this decision because it does not condition on the evolving learner state. The results suggest that adaptive teacher behavior is most valuable when the student simulator exposes enough cognitive detail for adaptation to matter.

\labeledlead{Backbone effect}{The underlying LLM substantially shapes emergent simulation behavior.}
GPT-5-based simulations generate larger knowledge graphs, reflected in high raw counts of both high- and low-mastery nodes. DeepSeek-V3.2 and Gemini-2.5-flash yield more moderate and internally consistent knowledge states. This supports the need to evaluate educational simulation frameworks across multiple backbone models rather than treating agent design as independent of model behavior.

The backbone effect also cautions against reporting a single aggregate score for an agent architecture. The same architecture may behave differently when implemented on a model that tends to elaborate concepts, a model that is conservative in knowledge-node generation, or a model that is prone to over-explanation. AgentSchool therefore treats the backbone model as part of the experimental condition. In future validation, calibration parameters should be estimated separately for different backbones, and robustness should be reported as cross-model stability rather than assumed from one implementation.

Overall, the lesson simulation indicates that structured agent design changes the nature of the generated evidence. Baseline student simulation tends to produce flatter outcome profiles, while AgentSchool produces a richer distribution of mastery, weakness, and misconception. Baseline or scripted teaching can produce plausible lessons, but adaptive teaching becomes more meaningful to evaluate when paired with a student model that exposes state transitions. These findings support the central claim that educational simulation should model learning as state change rather than as persona-conditioned dialogue.

\begin{remarkboxenv}{Lesson Simulation Takeaway}
$S_{\text{ours}}$ produces more differentiated and pedagogically inspectable student traces than the baseline simulator. Teacher-agent results are more conditional: $T_{\text{ours}}$ shows patterns consistent with ZPD-informed adaptation, but its advantage depends on backbone model and student-state observability.
\end{remarkboxenv}

\subsection{Social Simulation}
\guidingquestion{Can AgentSchool capture informal learning and classroom social dynamics beyond formal instruction?}

For recess socialization, we inspected three conditions over 300 interaction rounds: 11 agents, 11 agents with two aggressive agents, and a larger 30-agent setting. Agents autonomously chose whether to initiate, join, or exit topic discussions based on their interests and personalities. We represented relationships through social network structures and observed their evolution.

The social simulation differs from the lesson simulation in two ways. First, there is no explicit teacher-defined instructional objective; the scene is driven by peer interest, personality, and local social opportunity. Second, the main outcome is not concept mastery but network structure. We model the classroom as a weighted graph $G_t=(V,E_t,w_t)$, where each node is a student and each edge weight $w_{ij,t}$ represents the current relationship strength between students $i$ and $j$. Interaction events update edge weights, and the evolving graph is analyzed for isolation, clustering, centrality, and cohesion.

Several network indicators are useful for interpreting the generated dynamics. Degree centrality $d_i=\sum_j \mathbb{I}[w_{ij}>0]$ measures how many peers a student is connected to. Weighted strength $s_i=\sum_j w_{ij}$ measures the total quality of those connections. Clustering coefficient measures whether a student's friends are also connected to each other, indicating clique formation. Average path length and modularity indicate whether the class is integrated or segmented into subgroups. These metrics translate qualitative observations, such as ``popular student'' or ``isolated student,'' into inspectable evidence.

\begin{figure}[h]
    \centering
    \includegraphics[width=\textwidth]{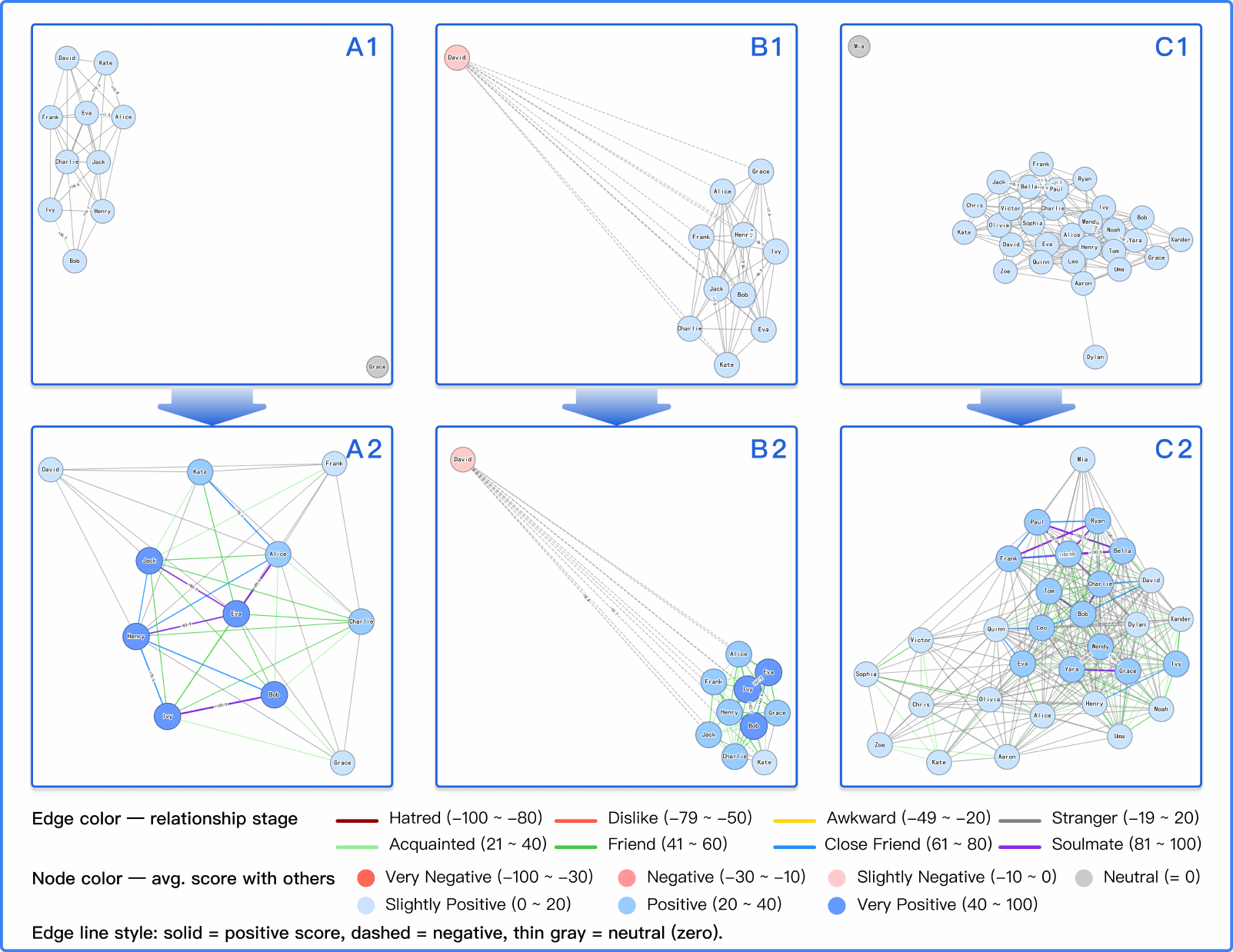}
    \caption{(A) The simulation with 11 agents. (B) The simulation with 11 agents in which two agents exhibited aggressiveness. (C) The simulation with 30 agents.}
    \label{fig:SN}
\end{figure}

\begin{table}[!htbp]
\centering
\caption{Qualitative social-simulation evidence and claim boundaries.}
\label{tab:social_evidence}
\renewcommand{\arraystretch}{1.08}
\begin{tabularx}{\linewidth}{@{}p{0.22\linewidth}X X@{}}
\toprule
\textbf{Observed phenomenon} & \textbf{Network evidence inspected} & \textbf{Responsible interpretation} \\
\midrule
Peripheral participation & Initially isolated or low-degree agents gradually formed weak ties and stabilized near the network periphery. & Supports phenomenon-level plausibility of gradual participation; does not estimate real-world prevalence. \\
Aggressive behavior & Edges around aggressive agents weakened, while ties among non-aggressive agents became denser in the same run. & Indicates second-order group effects in simulation; requires empirical social-network calibration before policy use. \\
Clique and hub formation & The larger setting produced dense local groups, dyadic ties, and visibly central agents. & Suggests the scenery can generate inspectable peer-influence structures; centrality values should be reported in future quantitative validation. \\
\bottomrule
\end{tabularx}
\end{table}

\labeledlead{Peripheral participation}{Introverted agents often began as isolated nodes, gradually contacted a few peers, and eventually stabilized near the group periphery.}
This trajectory resembles real classroom socialization, where peripheral membership can persist even after initial social contact.

The important point is that peripheral status is dynamic rather than preassigned. Introversion increases the probability of delayed participation, but it does not deterministically fix a student outside the group. Some agents first connect through low-risk topics, then gradually participate in broader conversation. Others remain weakly connected because their interests, timing, or emotional responses do not align with the dominant peer group. This supports the claim that AgentSchool can represent social participation as an unfolding process rather than as a static personality label.

\labeledlead{Aggressive behavior}{Aggressive agents became increasingly distant from the group, while non-aggressive agents became more cohesive.}
This suggests that the simulator can represent not only individual social outcomes, but also second-order group effects caused by disruptive behavior.

This second-order effect is educationally meaningful. In classrooms, disruptive behavior can isolate the aggressive student while simultaneously strengthening bonds among others through shared avoidance, mutual support, or collective norm formation. A simulator that only models pairwise dislike would miss this group-level consequence. AgentSchool's network representation captures both the direct edge weakening around aggressive agents and the indirect strengthening among non-aggressive agents.

\labeledlead{Clique and hub formation}{The 30-agent simulation produced small cliques, dyadic relationships, and central social hubs.}
Some agents moved into the inner circle of the classroom network, indicating stronger average relationship quality and potential opinion-leader roles.

The emergence of hubs matters because peer influence is often mediated by socially central students. If an AI tool, classroom reform, or learning norm is accepted by central students, it may diffuse more quickly; if central students resist, implementation may fail despite teacher effort. AgentSchool can therefore support experiments on how informal social networks interact with formal educational interventions. For example, a researcher can compare whether a collaborative learning design works differently when high-centrality students are supportive, indifferent, or resistant.

\remarkbox{These preliminary social simulations show that AgentSchool can generate inspectable traces of informal educational dynamics such as isolation, cohesion, clique formation, and peer influence. This makes it useful for hypothesis generation in pedagogy, classroom management, and campus governance, while quantitative deployment still requires empirical social-network calibration.}

\subsection{User Interface}

To support a broader range of educational stakeholders, we developed the AgentSchool web platform. The interface allows users to create simulation projects, configure agent roles, select scenery, and start simulations through UI-based interaction.

The interface is designed around the workflow of educational inquiry rather than around model engineering. A user first defines the simulation purpose, such as evaluating a lesson plan, comparing classroom layouts, or observing peer interaction. The user then configures agents and scenarios through interpretable parameters instead of editing raw prompts. After simulation, the platform presents both narrative traces and structured indicators, allowing users to inspect what happened, why the system inferred a particular outcome, and which state variables changed.


\begin{figure}[h]
    \centering
    \includegraphics[width=\textwidth]{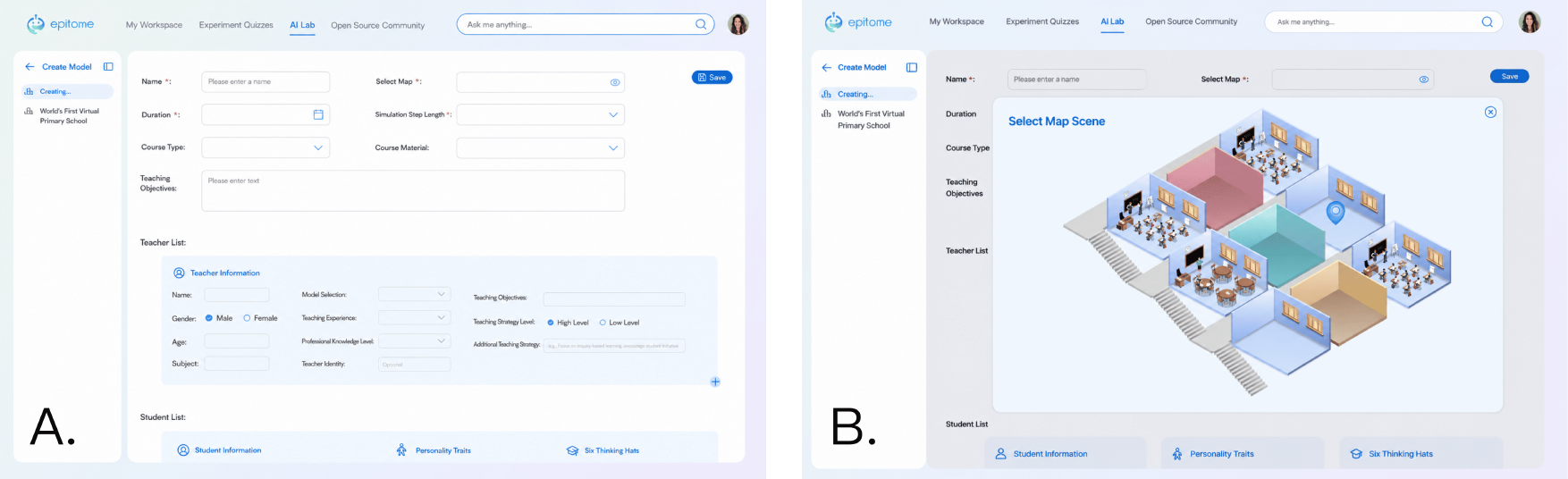}
    \caption{(A) Customized configuration webpage of one simulation. (B) Free choice among predefined scenery webpage}
    \label{fig:UIA}
\end{figure}

Figure~\ref{fig:UIA}A shows the configuration page, where users specify simulation metadata, class goals, instructional content, teacher-agent settings, and student-agent attributes such as personality, initial stress level, and misconceptions. Figure~\ref{fig:UIA}B shows the scenery selection page, including traditional and open classroom templates.

This configuration layer is important for reproducibility. Educational stakeholders often need to compare conditions while holding other variables constant. For example, a researcher may want to change only the teacher scaffold type while keeping student profiles and classroom layout fixed; a school administrator may want to compare traditional and open classrooms under the same lesson objective; a policymaker may want to examine how different grouping rules affect equity indicators. The UI stores these settings as project-level configurations, making it easier to rerun, audit, and share simulation conditions.

\begin{figure}[h]
    \centering
    \includegraphics[width=\textwidth]{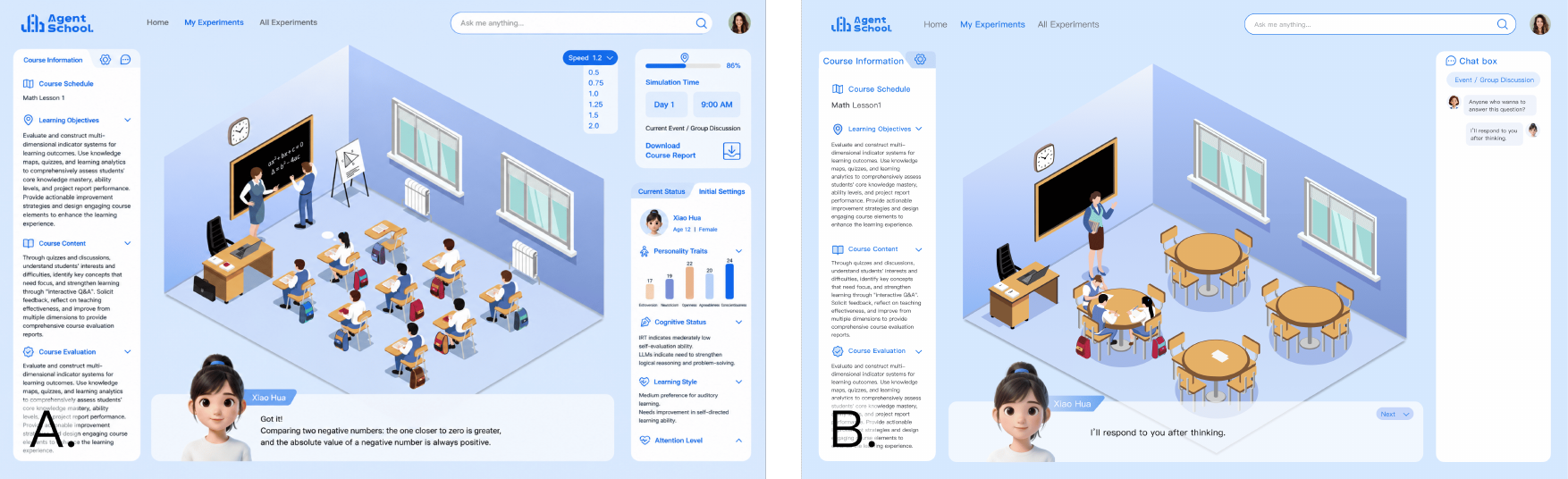}
    \caption{(A) The result webpage of a traditional classroom simulation.(B) The result webpage of a open classroom simulation.}
    \label{fig:UIB}
\end{figure}

Figure~\ref{fig:UIB}A presents a traditional classroom simulation result page. The main area displays the current scenery, the left panel shows lesson goals and content, and the lower-right panel presents the selected agent state. Users can adjust playback speed and download lesson evaluation results. Figure~\ref{fig:UIB}B presents an open classroom simulation, where the right panel records the conversation history of the study community.

The result page intentionally combines process playback with state inspection. A purely visual animation may be engaging but insufficient for research; a purely tabular output may be precise but difficult for educators to interpret. AgentSchool therefore connects event logs, dialogue history, agent state, and evaluation summaries. When a student answer changes, users can inspect the underlying knowledge node; when a group relation changes, users can inspect the interaction that caused it; when a teacher reflection is generated, users can inspect the evidence used. This design supports human-in-the-loop validation, where domain experts can judge whether generated trajectories are plausible and pedagogically meaningful.

\remarkbox{The UI makes AgentSchool usable for non-technical stakeholders, while the planned open-source framework will support deeper customization for researchers and developers.}

\section{Discussion}
\label{sec:discussion}
\sectionsubtitle{AgentSchool connects micro-level learning processes with meso-level classroom interaction, and defines a path toward macro-level institutional simulation.}

\labeledlead{Learning fields}{AgentSchool treats educational environments as configurable foundations for future modalities.}
Physical spaces, educational equipment, social networks, and interaction patterns jointly determine what kinds of learning can occur. By making these components explicit, AgentSchool allows stakeholders to prototype emerging educational paradigms before implementing them in real schools.

The concept of a learning field shifts attention from isolated instructional content to the conditions under which content becomes meaningful. A lesson about geography, for example, is not the same event when delivered through a teacher-centered lecture, a map-based group investigation, or a peer debate about regional development. The same knowledge point may be memorized, argued over, applied, or ignored depending on the field. This has practical significance for AI-enabled education: many proposed AI tools are evaluated as if they were independent interventions, but their effects depend on classroom norms, teacher roles, student relationships, and assessment pressure. AgentSchool provides a way to vary these contextual factors systematically.

This also changes how we think about future education. New educational scenarios should not be judged only by whether they resemble existing classrooms. If the purpose of simulation is to explore possible futures, then fidelity to the present is not always the correct objective. A future learning center with AI tutors, human mentors, flexible grouping, and project-based assessment may look unfamiliar, but it can still be evaluated by asking whether it supports coherent learning trajectories, equitable participation, safe social dynamics, and sustainable teacher work. AgentSchool's scenery abstraction is designed to make such questions operational.

\labeledlead{AI teachers}{Teacher agents support exploration of pedagogical pathways.}
Through flexible combinations of scaffolding strategies, planning modules, pedagogical principles, and reflective memory, AgentSchool can compare how different teaching styles interact with different student states and scenarios.

The teacher-agent design also provides a testbed for human-AI collaborative teaching. Rather than replacing teachers with a single autonomous model, AgentSchool is designed to simulate different divisions of labor: AI may generate diagnostic questions, human teachers may choose among scaffolds, classroom agents may monitor group dynamics, and future institutional agents may evaluate aggregate outcomes. Such configurations are difficult to test directly in schools because they change teacher workload, professional identity, and accountability. In simulation, they can be rehearsed and stress-tested. For example, researchers can ask whether an AI-generated lesson plan remains effective when student misconceptions deviate from expected patterns, or whether a teacher agent becomes overly dependent on automated diagnosis.

The results also suggest that adaptivity should be evaluated against learner state rather than against lesson-plan quality alone. A beautifully structured explanation can still fail if it addresses the wrong misconception or exceeds the student's current readiness. Conversely, a short hint may be powerful if it falls within the learner's ZPD. AgentSchool makes this distinction visible by linking teacher moves to subsequent changes in student knowledge graphs and misconception objects, while future validation should add action-level ZPD metrics.

\labeledlead{AI students}{Student agents make the learning process observable as a trajectory.}
Rather than measuring only final outputs, AgentSchool tracks cognitive state, knowledge-graph change, reasoning workflows, misconceptions, and social participation. This supports richer analysis of how learning outcomes emerge.

Student agents are not intended to replace empirical learners, and their outputs should not be treated as direct evidence about real students without calibration. Their value lies in enabling controlled exploration of mechanisms that are otherwise hard to observe. Real learning states are partially hidden; teachers infer them from answers, behavior, and assessment artifacts. AgentSchool makes the hidden state explicit so that researchers can compare internal trajectories with external performance. This creates a useful methodological bridge: the simulator can generate hypotheses about which observable behaviors indicate particular internal states, and these hypotheses can later be tested against classroom data.

The growable-student architecture also supports risk analysis. Educational AI systems may create failure modes that do not appear in short usability studies, such as overreliance on hints, shallow fluency, persistent misconception despite correct answers, or social stratification caused by differential access. By tracking state over time, AgentSchool can reveal such trajectories before deployment. This is especially important for minors and vulnerable learners, where trial-and-error innovation carries ethical cost.

\labeledlead{Limitations}{AgentSchool is an instrument for disciplined exploration, not an oracle.}
The current system still faces several limitations. First, LLM-generated behavior remains sensitive to backbone model, prompt structure, and calibration data. Second, many educational constructs, such as motivation, identity, and belonging, are only partially represented in the current student state. Third, preliminary validation is mainly phenomenological; stronger statistical alignment requires longitudinal empirical datasets. Fourth, raw node-count metrics are affected by generated graph size and should not be interpreted as normalized achievement rates without total-node reporting. Fifth, simulated outcomes may inherit biases from both the backbone models and the theoretical assumptions encoded in the simulator. These limitations do not invalidate the approach, but they define the boundary of responsible use. AgentSchool should be used to generate hypotheses, compare plausible mechanisms, and identify risks, then be paired with expert review and field validation before high-stakes decisions.

\begin{table}[!htbp]
\centering
\caption{Why AgentSchool is also an agent-research testbed.}
\label{tab:agent_testbed}
\renewcommand{\arraystretch}{1.08}
\begin{tabularx}{\linewidth}{@{}p{0.23\linewidth}X X@{}}
\toprule
\textbf{Agent challenge} & \textbf{Educational instantiation} & \textbf{Observable evidence} \\
\midrule
Long-horizon memory & Learners carry misconceptions, partial mastery, peer history, and teacher feedback across scenes. & Stability and revision of $M_{i,t}$, $K_{i,t}$, and $Z_{i,t}$ over simulated time. \\
Heterogeneous coordination & Teachers, students, and peer groups act under different roles, information, and goals. & Dialogue traces, interaction graphs, group participation, and scaffold selection. \\
Institutional reasoning & Future extensions add principals, evaluators, policy rules, and legitimacy pressure. & Counterfactual trajectories under changed schedules, assessment rules, or governance constraints. \\
Human-in-the-loop inspection & Educators need to inspect why a simulated outcome occurred before using it for decisions. & Linked logs connecting events, state updates, metrics, and replayable UI evidence. \\
\bottomrule
\end{tabularx}
\end{table}

\begin{remarkboxenv}{Broader Applicability}
Although AgentSchool is optimized for education, its simulator architecture can support broader multi-agent scenarios that require evolving agents, heterogeneous interaction patterns, and long-horizon coordination, such as collaborative work, organizational decision-making, and policy rehearsal.
\end{remarkboxenv}

\section{Future Work}
\label{sec:futurework}
\sectionsubtitle{Future development will expand AgentSchool from high-fidelity classroom simulation toward comprehensive school-system simulation across longer timescales and larger populations.}

\begin{figure}[h]
    \centering
    \includegraphics[width=\textwidth]{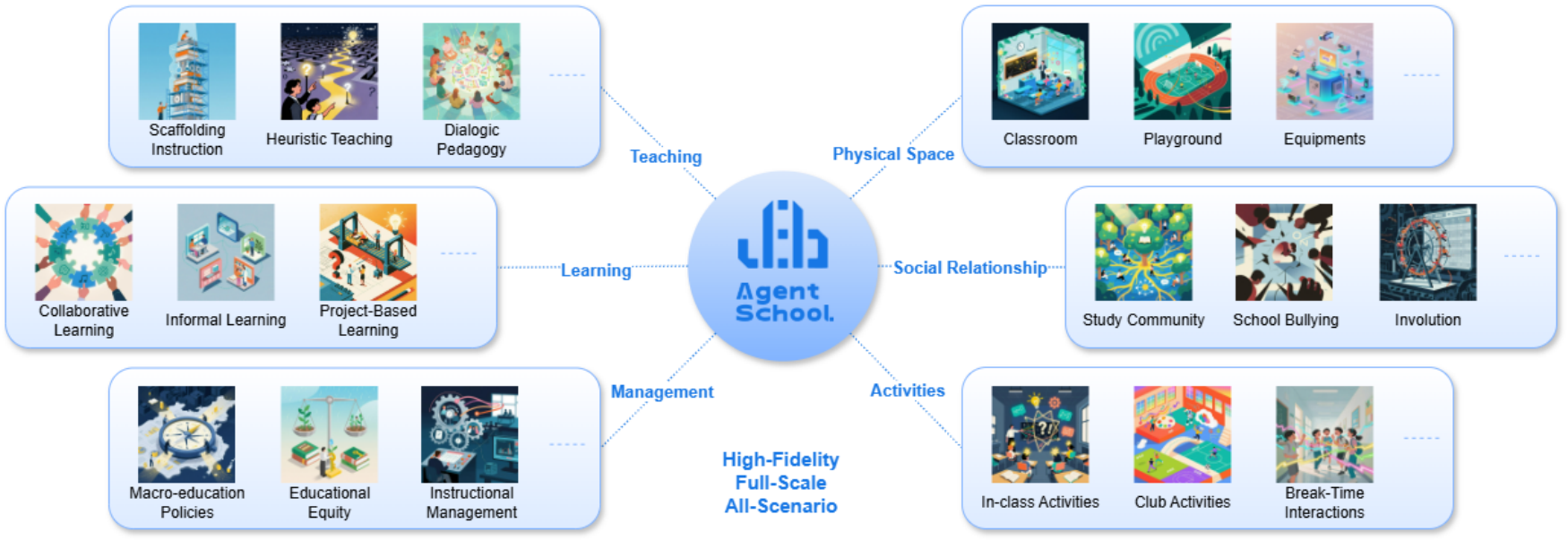}
    \caption{The envision of AgentSchool}
    \label{fig:vision_map}
\end{figure}

To better fulfill the role of a ``wind tunnel,'' AgentSchool will expand along five directions:

\begin{enumerate}[label=\textbf{F\arabic*.}, leftmargin=*]
    \item \textbf{Richer student agents.} We will extend student states from cognition to non-cognitive dimensions such as emotion, motivation, identity, values, and psychological development.
    \item \textbf{Richer learning fields.} We will model physical school spaces and facilities, including classrooms, laboratories, computer rooms, campus infrastructure, and teaching equipment. Agents will interact with these spaces to accumulate more diverse experiences.
    \item \textbf{Richer social and administrative scenarios.} We will incorporate student management, teacher management, educational project management, and policy implementation. This requires additional roles such as principals, curriculum coordinators, and policymakers.
    \item \textbf{Larger simulation scale.} We will expand from individuals and small classes to schools, cities, regions, and potentially national-level educational systems.
    \item \textbf{Longer simulation duration.} We will extend from single lessons and single days to months, years, and decade-level trajectories, with adaptive granularity to balance reliability and resource cost.
\end{enumerate}

\labeledlead{Richer student agents}{Future student models should represent the whole learner rather than only cognitive mastery.}
Knowledge growth is inseparable from motivation, emotion, self-efficacy, identity, and belonging. A student may understand a concept but avoid participation because of social anxiety; another may persist through difficulty because of strong interest or peer support. We therefore plan to extend the student state to include non-cognitive variables and their interaction with cognitive growth. This requires careful design because such constructs are ethically sensitive and difficult to validate. They should be represented as probabilistic, interpretable dimensions rather than as deterministic labels.

\labeledlead{Richer learning fields}{Future sceneries should include the material and institutional infrastructure of schools.}
Current AgentSchool sceneries focus on classrooms and recess social interaction. Future versions will include laboratories, libraries, computer rooms, maker spaces, counseling rooms, administrative offices, online platforms, and hybrid learning environments. These spaces are not merely locations; they encode access, authority, resources, and norms. Modeling them will allow researchers to study how school infrastructure shapes opportunity and how AI systems interact with physical and organizational design.

\labeledlead{Richer governance roles}{School systems require agents beyond students and teachers.}
Principals, grade leaders, curriculum coordinators, parents, policymakers, technology vendors, and evaluation bodies all influence educational change. Their actions often operate indirectly through incentives, schedules, assessment rules, procurement decisions, or professional norms. Adding these roles will allow AgentSchool to represent how a classroom innovation succeeds, diffuses, mutates, or fails within an organization. This is necessary for studying path dependency and institutional isomorphism, where schools may adopt new AI tools for legitimacy while preserving old routines.

\labeledlead{Larger scale}{Scaling should preserve educational meaning rather than only increasing agent count.}
Future work will extend AgentSchool from small classrooms to schools, districts, regions, and national systems. However, scale is useful only if the simulator preserves the mechanisms relevant to the research question. A million shallow agents are less informative than a smaller number of calibrated agents when the question concerns learning trajectories. We therefore plan to combine detailed micro-simulation for representative settings with aggregate or mean-field approximations for large populations, allowing the system to move between fidelity and breadth.

\labeledlead{Longer duration}{Long-horizon simulation must model both continuity and change.}
Education unfolds over years. Early misconceptions, teacher expectations, peer status, assessment experiences, and institutional decisions can accumulate into durable pathways. Future AgentSchool versions will support adaptive time compression so that critical moments are simulated in detail while routine periods are summarized through structured transitions. This will enable studies of delayed effects, such as whether an AI-supported assessment policy improves immediate feedback but gradually narrows curriculum, or whether a new collaborative learning design initially increases conflict but later improves classroom cohesion.

\labeledlead{Validation roadmap}{The next stage is systematic empirical calibration.}
The most important future task is to compare simulated trajectories with real educational data at multiple levels. At the micro level, simulated discourse can be compared with classroom interaction transcripts. At the learner level, mastery and misconception trajectories can be compared with repeated assessments and teacher annotations. At the social level, network evolution can be compared with sociometric or observational data. At the institutional level, policy simulations can be compared with administrative records and historical reform cases. Such validation will not produce a single final realism score; it will identify which mechanisms are reliable under which conditions.


\remarkbox{These enhancements will position AgentSchool as a comprehensive platform for exploring pedagogical innovation, school governance, educational policy, and institutional reform. Its long-term goal is to become a rigorous computational testbed for evidence-based educational transformation.}

\section{Conclusion}
\label{sec:conclusion}

This paper introduced AgentSchool, an LLM-powered multi-agent simulation platform for rehearsing future educational systems before they are deployed in real classrooms. The central argument is that educational AI cannot be validated only as a functional technology. Because it acts on developing learners, reshapes teacher practice, and interacts with institutional routines, it must be evaluated as part of a dynamic educational ecology. AgentSchool addresses this requirement by modeling students, teachers, scenarios, and interactions as evolving stateful systems rather than as static role descriptions.

The main methodological contribution is a shift from persona-conditioned role-play to structured educational state transition. Student agents maintain weighted knowledge graphs, thinking workflow pools, memories, learning parameters, and explicit misconception objects. Teacher agents plan, scaffold, diagnose, reflect, and update their experiential knowledge. Scenery objects define the relational field in which learning occurs, including formal classrooms, informal peer contexts, and future-oriented scenario candidates. The simulator coordinates these components across different scales, durations, and temporal granularities, making it possible to inspect both observable behavior and hidden state change.

Preliminary experiments show that AgentSchool produces richer and more interpretable dynamics than baseline role-play approaches within the tested settings. In lesson simulation, the proposed student agents generate more differentiated mastery profiles and misconception structures. The teacher-agent results are more backbone-dependent, but they show patterns consistent with ZPD-informed scaffolding when student-state information is available. In social simulation, the system generates plausible informal classroom phenomena, including peripheral participation, aggression-induced distance, group cohesion, clique formation, and opinion-leader emergence. These findings do not yet establish full statistical equivalence with real classrooms, but they support AgentSchool's value as a theory-grounded exploratory instrument.

Ultimately, AgentSchool is intended to serve as a computational wind tunnel for educational transformation. It allows researchers, educators, and policymakers to rehearse possible futures, examine failure modes, compare interventions, and generate hypotheses before real learners bear the cost of experimentation. Its responsible use requires calibration, expert review, and empirical validation, but its core promise is clear: by making educational trajectories simulatable and inspectable, we can reason about AI-enabled education with greater care, precision, and imagination.

\bibliographystyle{plainnat}
\bibliography{reference-whl}

@article{Mhlanga2024,
  author = {Jere, Samuel},
  title = {Evaluating Artificial Intelligence Large Language Models' Performances in a South African High School Chemistry Exam},
  journal = {Eurasia Journal of Mathematics, Science and Technology Education},
  year = {2025},
  volume = {21},
  number = {2},
  pages = {em2582},
  doi = {10.29333/ejmste/15932},
  url = {https://doi.org/10.29333/ejmste/15932}
}

@article{Chavda2023,
  author = {Chavda, Mrunal and Patel, Harsh and Bhatt, Hetav},
  title = {Quality Education through Writing: Aligning Learning Objectives in Learning Materials and Question Papers Using Bloom's Taxonomy},
  journal = {Quality Assurance in Education},
  year = {2023},
  volume = {32},
  number = {1},
  pages = {96--110},
  doi = {10.1108/QAE-03-2023-0045},
  url = {https://doi.org/10.1108/QAE-03-2023-0045}
}

@incollection{Lancaster2024,
  author = {Lancaster, Thomas and Clarke, Robert},
  title = {Contract Cheating: The Outsourcing of Assessed Student Work},
  booktitle = {Handbook of Academic Integrity},
  year = {2016},
  pages = {639--654},
  publisher = {Springer Singapore},
  address = {Singapore},
  doi = {10.1007/978-981-287-098-8_17},
  url = {https://doi.org/10.1007/978-981-287-098-8_17}
}

@article{Bozkurt2024,
  author = {Deep, Promethi Das and Chen, Yixin},
  title = {The Role of AI in Academic Writing: Impacts on Writing Skills, Critical Thinking, and Integrity in Higher Education},
  journal = {Societies},
  year = {2025},
  volume = {15},
  number = {9},
  pages = {247},
  doi = {10.3390/soc15090247},
  url = {https://doi.org/10.3390/soc15090247}
}

@incollection{Smith2024,
  author = {Smith, David and Francis, Nigel},
  title = {Process Not Product in the Written Assessment},
  booktitle = {Using Generative AI Effectively in Higher Education: Sustainable and Ethical Practices for Learning, Teaching and Assessment},
  year = {2024},
  publisher = {Routledge},
  address = {London},
  url = {https://www.routledge.com/Using-Generative-AI-Effectively-in-Higher-Education-Sustainable-and-Ethical-Practices-for-Learning-Teaching-and-Assessment/McDonald-Brown-Campus/p/book/9781032743912}
}

@article{Cela2024,
  author = {Zhao, Jian and Chapman, Elaine and Sabet, Peyman G. P.},
  title = {Generative AI and Educational Assessments: A Systematic Review},
  journal = {Education Research and Perspectives},
  year = {2024},
  volume = {51},
  pages = {124--155},
  doi = {10.70953/erpv51.2412006},
  url = {https://doi.org/10.70953/erpv51.2412006}
}

@article{Pallant2025,
  author = {Pallant, Jessica L. and Blijlevens, Janneke and Campbell, Alexander and Jopp, Ryan},
  title = {Mastering Knowledge: The Impact of Generative AI on Student Learning Outcomes},
  journal = {Studies in Higher Education},
  year = {2025},
  volume = {51},
  number = {4},
  pages = {714--735},
  doi = {10.1080/03075079.2025.2487570},
  url = {https://doi.org/10.1080/03075079.2025.2487570}
}

@article{Bearman2024,
  author = {Bearman, Margaret and Tai, Joanna and Dawson, Phillip and Boud, David and Ajjawi, Rola},
  title = {Developing Evaluative Judgement for a Time of Generative Artificial Intelligence},
  journal = {Assessment \& Evaluation in Higher Education},
  year = {2024},
  volume = {49},
  number = {6},
  pages = {893--905},
  doi = {10.1080/02602938.2024.2335321},
  url = {https://doi.org/10.1080/02602938.2024.2335321}
}

@inproceedings{Wang2021,
  author = {Wang, Shufang and Wang, Guorui and Chen, Xiaorong and Wang, Weijia and Ding, Xiaoqing},
  title = {A Review of Content Analysis on China Artificial Intelligence ({AI}) Education Policies},
  booktitle = {Proceedings of the 2021 International Conference on Artificial Intelligence and Education},
  year = {2021},
  pages = {395--399},
  publisher = {Atlantis Press},
  doi = {10.2991/assehr.k.210715.075},
  url = {https://doi.org/10.2991/assehr.k.210715.075}
}

@article{AlEmran2025,
  author = {Cotilla Conceicao, Jose Manuel and van der Stappen, Esther},
  title = {The Impact of AI on Inclusivity in Higher Education: A Rapid Review},
  journal = {Education Sciences},
  year = {2025},
  volume = {15},
  number = {9},
  pages = {1255},
  doi = {10.3390/educsci15091255},
  url = {https://doi.org/10.3390/educsci15091255}
}

@techreport{Jefferson2024,
  author = {Jefferson, Cassia},
  title = {Rethinking Education with Generative AI: Examining the Intersection of AI, Learning, and Digital Inequality in the UK's Education System},
  institution = {Digital Poverty Alliance},
  year = {2024},
  month = dec,
  url = {https://digitalpovertyalliance.org/}
}

@article{Haleem2024,
  author = {Baker, Ryan S. and Hawn, Aaron},
  title = {Algorithmic Bias in Education},
  journal = {International Journal of Artificial Intelligence in Education},
  year = {2022},
  volume = {32},
  number = {4},
  pages = {1052--1092},
  doi = {10.1007/s40593-021-00285-9},
  url = {https://doi.org/10.1007/s40593-021-00285-9}
}

@misc{Bozkurt2024b,
  author = {Yin, Zhipeng and Chinta, Sribala Vidyadhari and Wang, Zichong and Gonzalez, Matthew and Zhang, Wenbin},
  title = {{FairAIED}: Navigating Fairness, Bias, and Ethics in Educational AI Applications},
  year = {2024},
  publisher = {arXiv},
  eprint = {2407.18745},
  archivePrefix = {arXiv},
  primaryClass = {cs.LG},
  doi = {10.48550/arXiv.2407.18745},
  url = {https://arxiv.org/abs/2407.18745}
}

@article{Floridi2018,
  author = {Floridi, Luciano and Cowls, Josh and Beltrametti, Monica and Chatila, Raja and Chazerand, Patrice and Dignum, Virginia and Luetge, Christoph and Madelin, Robert and Pagallo, Ugo and Rossi, Francesca and Schafer, Burkhard and Valcke, Peggy and Vayena, Effy},
  title = {{AI4People}: An Ethical Framework for a Good AI Society: Opportunities, Risks, Principles, and Recommendations},
  journal = {Minds and Machines},
  year = {2018},
  volume = {28},
  number = {4},
  pages = {689--707},
  doi = {10.1007/s11023-018-9482-5},
  url = {https://doi.org/10.1007/s11023-018-9482-5}
}

@article{Hagood2021,
  author = {Hagood, Dylan},
  title = {Standardized Testing: A History and Analysis},
  journal = {OUR Journal: ODU Undergraduate Research Journal},
  year = {2021},
  volume = {8},
  number = {1},
  url = {https://digitalcommons.odu.edu/ourj/vol8/iss1/4/}
}

@techreport{Champion2016,
  author = {Champion, Elizabeth and Barab, Sasha and Dodge, Tyler and Tucker-Raymond, Eli and Wohlwend, Karen and Maina, Faith},
  title = {Investigating the Affordances and Constraints of Digital Educational Games for Early Primary School Classrooms},
  institution = {STELAR},
  year = {2016},
  url = {https://stelar.edc.org/}
}

@article{Muvunyi2024,
  author = {Muvunyi, Emmanuel},
  title = {Aligning Budgets with Local Needs: Structural Barriers in Rwanda's Education Decentralisation},
  journal = {Journal of Research, Innovation, and Interdisciplinary Studies},
  year = {2024}
}

@article{Roza2010,
  author = {Roza, Marguerite},
  title = {Student-Based Budgeting: A Potentially Powerful Tool for Equity},
  journal = {Voices in Urban Education},
  year = {2010},
  number = {29},
  pages = {18--27}
}

@incollection{Stigler2015,
  author = {Hiebert, James},
  title = {The Constantly Underestimated Challenge of Improving Mathematics Instruction},
  booktitle = {Vital Directions for Mathematics Education Research},
  year = {2013},
  pages = {45--56},
  publisher = {Springer New York},
  address = {New York, NY},
  doi = {10.1007/978-1-4614-6977-3_3},
  url = {https://doi.org/10.1007/978-1-4614-6977-3_3}
}

@techreport{Doering2002,
  author = {Doering, Aaron},
  title = {Assisting Teachers in Unlearning Their Standard Organization's Operating Procedures: A Case Study},
  institution = {ERIC},
  year = {2002},
  number = {ED462952},
  url = {https://eric.ed.gov/?id=ED462952}
}

@techreport{Sotiriou2018,
  author = {Sotiriou, Sofoklis A. and Bogner, Franz X. and Hohenwarter, Markus and L{'o}pez, Ver{'o}nica and Arnal, Marta and De Vocht, Michiel},
  title = {D2.1 Open Schooling Model},
  institution = {Open Schools for Open Societies},
  year = {2018},
  url = {https://www.openschools.eu/}
}

@article{Tulyakul2023,
  author = {Tulyakul, Sarin},
  title = {A Critical Review of Educational Innovation: Challenges, Critiques, and Policy Suggestions},
  journal = {ASEAN Journal of Health and Social Issues},
  year = {2023},
  volume = {4},
  number = {1}
}

@inproceedings{Park2023,
  author = {Park, Joon Sung and O'Brien, Joseph C. and Cai, Carrie J. and Morris, Meredith Ringel and Liang, Percy and Bernstein, Michael S.},
  title = {Generative Agents: Interactive Simulacra of Human Behavior},
  booktitle = {Proceedings of the 36th Annual ACM Symposium on User Interface Software and Technology},
  year = {2023},
  series = {UIST '23},
  pages = {1--22},
  publisher = {Association for Computing Machinery},
  address = {New York, NY, USA},
  doi = {10.1145/3586183.3606763},
  url = {https://doi.org/10.1145/3586183.3606763}
}

@misc{An2021,
  author = {Ju, Da and Williams, Adina and Karrer, Brian and Nickel, Maximilian},
  title = {Sense and Sensitivity: Evaluating the Simulation of Social Dynamics via Large Language Models},
  year = {2024},
  eprint = {2412.05093},
  archivePrefix = {arXiv},
  primaryClass = {cs.CL},
  doi = {10.48550/arXiv.2412.05093},
  url = {https://arxiv.org/abs/2412.05093}
}

@misc{piao2025agentsociety,
  author = {Piao, Jinghua and Yan, Yuwei and Zhang, Jun and Li, Nian and Yan, Junbo and Lan, Xiaochong and Lu, Zhihong and Zheng, Zhiheng and Wang, Jing Yi and Zhou, Di and Gao, Chen and Xu, Fengli and Zhang, Fang and Rong, Ke and Su, Jun and Li, Yong},
  title = {{AgentSociety}: Large-Scale Simulation of LLM-Driven Generative Agents Advances Understanding of Human Behaviors and Society},
  year = {2025},
  eprint = {2502.08691},
  archivePrefix = {arXiv},
  primaryClass = {cs.CY},
  doi = {10.48550/arXiv.2502.08691},
  url = {https://arxiv.org/abs/2502.08691}
}

@inproceedings{Zhang2025,
  author = {Zhang, Zheyuan and Zhang-Li, Daniel and Yu, Jifan and Gong, Linlu and Zhou, Jinchang and Hao, Zhanxin and Jiang, Jianxiao and Cao, Jie and Liu, Huiqin and Liu, Zhiyuan and Hou, Lei and Li, Juanzi},
  title = {Simulating Classroom Education with {LLM-Empowered} Agents},
  booktitle = {Proceedings of the 2025 Conference of the Nations of the Americas Chapter of the Association for Computational Linguistics: Human Language Technologies},
  year = {2025},
  publisher = {Association for Computational Linguistics},
  url = {https://arxiv.org/abs/2406.19226},
  eprint = {2406.19226},
  archivePrefix = {arXiv},
  primaryClass = {cs.CL}
}

@misc{Anon2024_PolicySim,
  author = {Liu, Lu and Liu, Zhimin},
  title = {The Paradigm Shift of Sustainable Development in Educational Economics in the Digitization-Intelligent Era: A Kuhnian Analysis},
  year = {2025},
  publisher = {Preprints.org},
  doi = {10.20944/preprints202509.1463.v1},
  url = {https://doi.org/10.20944/preprints202509.1463.v1}
}

@inproceedings{Zhao2024d,
  author = {Zhang, Zheyuan and Zhang-Li, Daniel and Yu, Jifan and Gong, Linlu and Zhou, Jinchang and Hao, Zhanxin and Jiang, Jianxiao and Cao, Jie and Liu, Huiqin and Liu, Zhiyuan and Hou, Lei and Li, Juanzi},
  title = {Simulating Classroom Education with {LLM-Empowered} Agents},
  booktitle = {Proceedings of the 2025 Conference of the Nations of the Americas Chapter of the Association for Computational Linguistics: Human Language Technologies},
  year = {2025},
  publisher = {Association for Computational Linguistics},
  url = {https://arxiv.org/abs/2406.19226},
  eprint = {2406.19226},
  archivePrefix = {arXiv},
  primaryClass = {cs.CL}
}

@article{DiMaggio1983,
  title = {The Iron Cage Revisited: Institutional Isomorphism and Collective Rationality in Organizational Fields},
  author = {DiMaggio, Paul J. and Powell, Walter W.},
  journal = {American Sociological Review},
  volume = {48},
  number = {2},
  pages = {147--160},
  year = {1983},
  doi = {10.2307/2095101}
}

@article{garcia-magarino_ftssociagentbasedframework_2015,
  author = {Garcia-Magarino, Ivan and Plaza, Inmaculada},
  title = {{FTS-SOCI}: An Agent-Based Framework for Simulating Teaching Strategies with Evolutions of Sociograms},
  journal = {Simulation Modelling Practice and Theory},
  year = {2015},
  volume = {57},
  pages = {161--178},
  doi = {10.1016/j.simpat.2015.07.003},
  url = {https://doi.org/10.1016/j.simpat.2015.07.003}
}

@misc{park_generativeagentsinteractive_2023,
  author = {Park, Joon Sung and O'Brien, Joseph C. and Cai, Carrie J. and Morris, Meredith Ringel and Liang, Percy and Bernstein, Michael S.},
  title = {Generative Agents: Interactive Simulacra of Human Behavior},
  year = {2023},
  eprint = {2304.03442},
  archivePrefix = {arXiv},
  primaryClass = {cs.HC},
  doi = {10.48550/arXiv.2304.03442},
  url = {https://arxiv.org/abs/2304.03442}
}

@misc{hicke_medsimaisimulationformative_2025,
  author = {Hicke, Yann and Geathers, Jadon and Rajashekar, Niroop and Chan, Colleen and Jack, Anyanate Gwendolyne and Sewell, Justin and Preston, Mackenzi and Cornes, Susannah and Shung, Dennis and Kizilcec, Rene},
  title = {{MedSimAI}: Simulation and Formative Feedback Generation to Enhance Deliberate Practice in Medical Education},
  year = {2025},
  eprint = {2503.05793},
  archivePrefix = {arXiv},
  primaryClass = {cs.CY},
  doi = {10.48550/arXiv.2503.05793},
  url = {https://arxiv.org/abs/2503.05793}
}

@misc{guan_modelingearthscalehumanlike_2025,
  author = {Guan, Haoxiang and He, Jiyan and Fan, Liyang and Ren, Zhenzhen and He, Shaobin and Yu, Xin and Chen, Yuan and Zheng, Shuxin and Liu, Tie-Yan and Liu, Zhen},
  title = {Modeling Earth-Scale Human-like Societies with One Billion Agents},
  year = {2025},
  eprint = {2506.12078},
  archivePrefix = {arXiv},
  primaryClass = {cs.CY},
  doi = {10.48550/arXiv.2506.12078},
  url = {https://arxiv.org/abs/2506.12078}
}

@misc{piao_agentsocietylargescalesimulation_2025,
  author = {Piao, Jinghua and Yan, Yuwei and Zhang, Jun and Li, Nian and Yan, Junbo and Lan, Xiaochong and Lu, Zhihong and Zheng, Zhiheng and Wang, Jing Yi and Zhou, Di and Gao, Chen and Xu, Fengli and Zhang, Fang and Rong, Ke and Su, Jun and Li, Yong},
  title = {{AgentSociety}: Large-Scale Simulation of LLM-Driven Generative Agents Advances Understanding of Human Behaviors and Society},
  year = {2025},
  eprint = {2502.08691},
  archivePrefix = {arXiv},
  primaryClass = {cs.CY},
  doi = {10.48550/arXiv.2502.08691},
  url = {https://arxiv.org/abs/2502.08691}
}

@misc{yuan_simulatinghumanlikelearning_2025,
  author = {Yuan, Yu and Zhao, Lili and Chen, Wei and Zheng, Guangting and Zhang, Kai and Zhang, Mengdi and Liu, Qi},
  title = {Simulating Human-like Learning Dynamics with LLM-Empowered Agents},
  year = {2025},
  eprint = {2508.05622},
  archivePrefix = {arXiv},
  primaryClass = {cs.CY},
  doi = {10.48550/arXiv.2508.05622},
  url = {https://arxiv.org/abs/2508.05622}
}

@misc{xie_canlargelanguage_2024,
  author = {Xie, Chengxing and Chen, Canyu and Jia, Feiran and Ye, Ziyu and Lai, Shiyang and Shu, Kai and Gu, Jindong and Bibi, Adel and Hu, Ziniu and Jurgens, David and Evans, James and Torr, Philip and Ghanem, Bernard and Li, Guohao},
  title = {Can Large Language Model Agents Simulate Human Trust Behavior?},
  year = {2024},
  eprint = {2402.04559},
  archivePrefix = {arXiv},
  primaryClass = {cs.AI},
  doi = {10.48550/arXiv.2402.04559},
  url = {https://arxiv.org/abs/2402.04559}
}

@misc{mi_mfllmsimulatingpopulation_2025,
  author = {Mi, Qirui and Yang, Mengyue and Yu, Xiangning and Zhao, Zhiyu and Deng, Cheng and An, Bo and Zhang, Haifeng and Chen, Xu and Wang, Jun},
  title = {{MF-LLM}: Simulating Population Decision Dynamics via a Mean-Field Large Language Model Framework},
  year = {2025},
  eprint = {2504.21582},
  archivePrefix = {arXiv},
  primaryClass = {cs.MA},
  doi = {10.48550/arXiv.2504.21582},
  url = {https://arxiv.org/abs/2504.21582}
}

@inproceedings{xu_classroomsimulacrabuilding_2025,
  author = {Xu, Songlin and Wen, Hao-Ning and Pan, Hongyi and Dominguez, Dallas and Hu, Dongyin and Zhang, Xinyu},
  title = {Classroom Simulacra: Building Contextual Student Generative Agents in Online Education for Learning Behavioral Simulation},
  booktitle = {Proceedings of the 2025 CHI Conference on Human Factors in Computing Systems},
  year = {2025},
  series = {CHI '25},
  pages = {1--26},
  publisher = {Association for Computing Machinery},
  address = {New York, NY, USA},
  doi = {10.1145/3706598.3713773},
  url = {https://doi.org/10.1145/3706598.3713773}
}

@misc{li_agenthospitalsimulacrum_2025,
  author = {Li, Junkai and Lai, Yunghwei and Li, Weitao and Ren, Jingyi and Zhang, Meng and Kang, Xinhui and Wang, Siyu and Li, Peng and Zhang, Ya-Qin and Ma, Weizhi and Liu, Yang},
  title = {Agent Hospital: A Simulacrum of Hospital with Evolvable Medical Agents},
  year = {2025},
  eprint = {2405.02957},
  archivePrefix = {arXiv},
  primaryClass = {cs.AI},
  doi = {10.48550/arXiv.2405.02957},
  url = {https://arxiv.org/abs/2405.02957}
}

@article{hu_exploringpotentialllm_2025,
  author = {Hu, Bihao and Zhu, Jiayi and Pei, Yiying and Gu, Xiaoqing},
  title = {Exploring the Potential of LLM to Enhance Teaching Plans through Teaching Simulation},
  journal = {npj Science of Learning},
  year = {2025},
  volume = {10},
  number = {1},
  pages = {1--12},
  doi = {10.1038/s41539-025-00300-x},
  url = {https://doi.org/10.1038/s41539-025-00300-x}
}

@techreport{MEC_smarteducationwhitebook,
  author = {{Ministry of Education of the People's Republic of China}},
  title = {White Paper on Smart Education in China},
  institution = {Ministry of Education of the People's Republic of China},
  year = {2025},
  month = may,
  address = {Beijing, China},
  url = {https://www.moe.gov.cn/jyb_xwfb/xw_zt/moe_357/2025/2025_zt06/cgfb/202505/t20250507_1189603.html}
}

@incollection{dewey2020education,
  author = {Dewey, John},
  title = {Education as Growth},
  booktitle = {Pragmatism},
  year = {2020},
  pages = {93--101},
  publisher = {Routledge},
  address = {London}
}

@article{basu_scaffoldingframeworksupport_2015,
  author = {Basu, Satabdi and Sengupta, Pratim and Biswas, Gautam},
  title = {A Scaffolding Framework to Support Learning of Emergent Phenomena Using Multi-Agent-Based Simulation Environments},
  journal = {Research in Science Education},
  year = {2015},
  volume = {45},
  number = {2},
  pages = {293--324},
  doi = {10.1007/s11165-014-9424-z},
  url = {https://doi.org/10.1007/s11165-014-9424-z}
}

@article{gu_systematicreviewagentbased_2015,
  author = {Gu, X. and Blackmore, K. L.},
  title = {A Systematic Review of Agent-Based Modelling and Simulation Applications in the Higher Education Domain},
  journal = {Higher Education Research \& Development},
  year = {2015},
  volume = {34},
  number = {5},
  pages = {883--898},
  doi = {10.1080/07294360.2015.1011088},
  url = {https://doi.org/10.1080/07294360.2015.1011088}
}

@article{bodine_agentbasedmodelingsimulation_2020,
  author = {Bodine, Erin N. and Panoff, Robert M. and Voit, Eberhard O. and Weisstein, Anton E.},
  title = {Agent-Based Modeling and Simulation in Mathematics and Biology Education},
  journal = {Bulletin of Mathematical Biology},
  year = {2020},
  volume = {82},
  number = {8},
  pages = {101},
  doi = {10.1007/s11538-020-00778-z},
  url = {https://doi.org/10.1007/s11538-020-00778-z}
}

@article{stummer_agentbasedmarketsimulation_2021,
  author = {Stummer, Christian and Kiesling, Elmar},
  title = {An Agent-Based Market Simulation for Enriching Innovation Management Education},
  journal = {Central European Journal of Operations Research},
  year = {2021},
  volume = {29},
  number = {1},
  pages = {143--161},
  doi = {10.1007/s10100-020-00716-3},
  url = {https://doi.org/10.1007/s10100-020-00716-3}
}

@article{dubovi_instructionalsupportlearning_2019,
  author = {Dubovi, Ilana and Lee, Victor R.},
  title = {Instructional Support for Learning with Agent-Based Simulations: A Tale of Vicarious and Guided Exploration Learning Approaches},
  journal = {Computers \& Education},
  year = {2019},
  volume = {142},
  pages = {103644},
  doi = {10.1016/j.compedu.2019.103644},
  url = {https://doi.org/10.1016/j.compedu.2019.103644}
}

@incollection{bhowmik_evaluationllmpoweredstudent_2024,
  author = {Bhowmik, Saptarshi and West, Luke and Barrett, Alex and Zhang, Nuodi and Dai, Chih-Pu and Sokolikj, Zlatko and Southerland, Sherry and Yuan, Xin and Ke, Fengfeng},
  title = {Evaluation of an LLM-Powered Student Agent for Teacher Training},
  booktitle = {Technology Enhanced Learning for Inclusive and Equitable Quality Education},
  year = {2024},
  volume = {15160},
  pages = {68--74},
  publisher = {Springer Nature Switzerland},
  address = {Cham},
  doi = {10.1007/978-3-031-72312-4_7},
  url = {https://doi.org/10.1007/978-3-031-72312-4_7}
}

@inproceedings{pan_tutorupwhatif_2025,
  author = {Pan, Sitong and Schmucker, Robin and Garcia Bulle Bueno, Bernardo and Llanes, Salome Aguilar and Albo Alarcon, Fernanda and Zhu, Hangxiao and Teo, Adam and Xia, Meng},
  title = {{TutorUp}: What If Your Students Were Simulated? Training Tutors to Address Engagement Challenges in Online Learning},
  booktitle = {Proceedings of the 2025 CHI Conference on Human Factors in Computing Systems},
  year = {2025},
  series = {CHI '25},
  pages = {1--18},
  publisher = {Association for Computing Machinery},
  address = {New York, NY, USA},
  doi = {10.1145/3706598.3713589},
  url = {https://doi.org/10.1145/3706598.3713589}
}

@inproceedings{takahashi_datascienceagentbased_2023,
  author = {Takahashi, Satoshi and Yoshikawa, Atsushi},
  title = {Data Science in an Agent-Based Simulation World},
  booktitle = {2023 IEEE International Conference on Teaching, Assessment and Learning for Engineering (TALE)},
  year = {2023},
  pages = {1--6},
  doi = {10.1109/TALE56641.2023.10398326},
  url = {https://doi.org/10.1109/TALE56641.2023.10398326}
}

@incollection{emond_cognitivesimulationsadaptive_2023,
  author = {Emond, Bruno and Zeinali-Torbati, Reza and Smith, Jennifer and Billard, Randy and Barnes, Joshua and Veitch, Brian},
  title = {Cognitive Simulations for Adaptive Instructional Systems: Exploring Instruction Strategies with Simulated Tutors and Learners},
  booktitle = {Adaptive Instructional Systems},
  year = {2023},
  pages = {123--136},
  publisher = {Springer Nature Switzerland},
  address = {Cham},
  doi = {10.1007/978-3-031-34735-1_9},
  url = {https://doi.org/10.1007/978-3-031-34735-1_9}
}

@incollection{alharbi_agentbasedclassroomenvironment_2021,
  author = {Alharbi, Khulood and Cristea, Alexandra I. and Shi, Lei and Tymms, Peter and Brown, Chris},
  title = {Agent-Based Classroom Environment Simulation: The Effect of Disruptive Schoolchildren's Behaviour versus Teacher Control over Neighbours},
  booktitle = {Artificial Intelligence in Education},
  year = {2021},
  pages = {48--53},
  publisher = {Springer International Publishing},
  address = {Cham},
  doi = {10.1007/978-3-030-78270-2_8},
  url = {https://doi.org/10.1007/978-3-030-78270-2_8}
}

@misc{rao_multiagentsystemcomprehensive_2025,
  author = {Rao, Jiayuan and Li, Zifeng and Wu, Haoning and Zhang, Ya and Wang, Yanfeng and Xie, Weidi},
  title = {Multi-Agent System for Comprehensive Soccer Understanding},
  year = {2025},
  eprint = {2505.03735},
  archivePrefix = {arXiv},
  primaryClass = {cs.CV},
  doi = {10.48550/arXiv.2505.03735},
  url = {https://arxiv.org/abs/2505.03735}
}

@article{nussbaum_alternativeframeworksconceptual_1982,
  author = {Nussbaum, Joseph and Novick, Shimshon},
  title = {Alternative Frameworks, Conceptual Conflict and Accommodation: Toward a Principled Teaching Strategy},
  journal = {Instructional Science},
  year = {1982},
  volume = {11},
  number = {3},
  pages = {183--200},
  doi = {10.1007/BF00414279},
  url = {https://doi.org/10.1007/BF00414279}
}

@techreport{europeancommission_proposalcouncilrecommendation_2018,
  author = {{European Commission}},
  title = {Proposal for a Council Recommendation on Key Competences for Lifelong Learning},
  institution = {European Commission},
  year = {2018},
  number = {COM(2018) 24 final},
  url = {https://eur-lex.europa.eu/legal-content/EN/TXT/?uri=CELEX:52018DC0024}
}

@misc{ieeesa_competenciesrequestinterest_2015,
  author = {{IEEE Standards Association}},
  title = {Competencies: Request for Interest},
  year = {2015},
  month = mar,
  url = {https://ieee-sa.imeetcentral.com/ltsc/doc/WzIsMzc3MzUxMDld/w-CompetenciesStudyGroupRequestForInterest}
}

@techreport{oecd_21stcenturyskills_2009,
  author = {Ananiadou, Katerina and Claro, Magdalean},
  title = {21st Century Skills and Competences for New Millennium Learners in OECD Countries},
  institution = {OECD Publishing},
  year = {2009},
  type = {OECD Education Working Papers},
  number = {41},
  doi = {10.1787/218525261154},
  url = {https://doi.org/10.1787/218525261154}
}

@article{annelinAssessmentKeySustainability2022,
  author = {Annelin, Alice and Bostrom, Gert-Olof},
  title = {An Assessment of Key Sustainability Competencies: A Review of Scales and Propositions for Validation},
  journal = {International Journal of Sustainability in Higher Education},
  year = {2022},
  volume = {24},
  number = {9},
  pages = {53--69},
  doi = {10.1108/IJSHE-05-2022-0166},
  url = {https://doi.org/10.1108/IJSHE-05-2022-0166}
}

@book{vygotskyumstvennoie1935,
  author = {Vygotskii, L. S.},
  title = {Umstvennoie Razvitie Detei v Protsesse Obuchenia},
  year = {1935},
  publisher = {Gosudarstvennoie Uchebno-pedagogicheskoie Izdatel'stvo},
  address = {Moscow},
  language = {Russian}
}

@article{wood_roletutoringproblem_1976,
  author = {Wood, David and Bruner, Jerome S. and Ross, Gail},
  title = {The Role of Tutoring in Problem Solving},
  journal = {Journal of Child Psychology and Psychiatry},
  year = {1976},
  volume = {17},
  number = {2},
  pages = {89--100},
  doi = {10.1111/j.1469-7610.1976.tb00381.x},
  url = {https://doi.org/10.1111/j.1469-7610.1976.tb00381.x}
}

@book{fischer_internationalhandbooklearning_2018,
  editor = {Fischer, Frank and Hmelo-Silver, Cindy E. and Goldman, Susan R. and Reimann, Peter},
  title = {International Handbook of the Learning Sciences},
  year = {2018},
  edition = {1},
  publisher = {Routledge},
  address = {New York, NY},
  doi = {10.4324/9781315617572},
  url = {https://doi.org/10.4324/9781315617572}
}

@article{barr_learninglessonssocial_1980,
  author = {Barr, Rebecca},
  title = {Learning Lessons: Social Organization in the Classroom. Hugh Mehan},
  journal = {American Journal of Education},
  year = {1980},
  volume = {88},
  number = {3},
  pages = {366--374},
  doi = {10.1086/443532},
  url = {https://doi.org/10.1086/443532}
}

@article{dillenbourg_mechanicscsclmacro_2008,
  author = {Dillenbourg, Pierre and Hong, Fabrice},
  title = {The Mechanics of CSCL Macro Scripts},
  journal = {International Journal of Computer-Supported Collaborative Learning},
  year = {2008},
  volume = {3},
  number = {1},
  pages = {5--23},
  doi = {10.1007/s11412-007-9033-1},
  url = {https://doi.org/10.1007/s11412-007-9033-1}
}

@article{tsovaltzi_groupawarenesssupport_2014,
  author = {Tsovaltzi, Dimitra and Puhl, Thomas and Judele, Raluca and Weinberger, Armin},
  title = {Group Awareness Support and Argumentation Scripts for Individual Preparation of Arguments in Facebook},
  journal = {Computers \& Education},
  year = {2014},
  volume = {76},
  pages = {108--118},
  doi = {10.1016/j.compedu.2014.03.012},
  url = {https://doi.org/10.1016/j.compedu.2014.03.012}
}

@article{gielen_scriptingroleassessor_2015,
  author = {Gielen, Mario and De Wever, Bram},
  title = {Scripting the Role of Assessor and Assessee in Peer Assessment in a Wiki Environment: Impact on Peer Feedback Quality and Product Improvement},
  journal = {Computers \& Education},
  year = {2015},
  volume = {88},
  pages = {370--386},
  doi = {10.1016/j.compedu.2015.07.012},
  url = {https://doi.org/10.1016/j.compedu.2015.07.012}
}

@book{thornburg_campfireholodeckcreating_2013,
  author = {Thornburg, David},
  title = {From the Campfire to the Holodeck: Creating Engaging and Powerful 21st Century Learning Environments},
  year = {2013},
  publisher = {John Wiley \& Sons},
  address = {Newark, NJ},
  isbn = {9781118748060}
}

@article{maroulis_modelingtransitionpublic_2014,
  author = {Maroulis, Spiro and Bakshy, Eytan and Gomez, Louis and Wilensky, Uri},
  title = {Modeling the Transition to Public School Choice},
  journal = {Journal of Artificial Societies and Social Simulation},
  year = {2014},
  volume = {17},
  number = {2},
  doi = {10.18564/jasss.2402},
  url = {https://doi.org/10.18564/jasss.2402}
}

@article{maroulis_complexsystemsview_2010,
  author = {Maroulis, S. and Guimera, R. and Petry, H. and Stringer, M. J. and Gomez, L. M. and Amaral, L. A. N. and Wilensky, U.},
  title = {Complex Systems View of Educational Policy Research},
  journal = {Science},
  year = {2010},
  volume = {330},
  number = {6000},
  pages = {38--39},
  doi = {10.1126/science.1195153},
  url = {https://doi.org/10.1126/science.1195153}
}

@misc{jiang_evolutionaryreinforcementlearning_2025,
  author = {Jiang, Mei and Shen, Haihai and Luo, Zhuo and Li, Bingdong and Hong, Wenjing and Tang, Ke and Zhou, Aimin},
  title = {Evolutionary Reinforcement Learning Based AI Tutor for Socratic Interdisciplinary Instruction},
  year = {2025},
  eprint = {2512.11930},
  archivePrefix = {arXiv},
  primaryClass = {cs.CY},
  doi = {10.48550/arXiv.2512.11930},
  url = {https://arxiv.org/abs/2512.11930}
}

\end{document}